\documentclass{article}
\usepackage{microtype}
\usepackage{graphicx}
\usepackage{subcaption}
\usepackage{booktabs}
\usepackage{hyperref}
\usepackage{multirow}
\usepackage{xcolor}
\usepackage{tabularx}
\usepackage{tikz}
\usetikzlibrary{arrows.meta,positioning,fit,calc}

\usepackage{algorithm}
\usepackage{algorithmic}
\usepackage[preprint]{icml2026}
\usepackage{amsmath}
\usepackage{amssymb}
\usepackage{mathtools}
\usepackage{amsthm}
\usepackage[capitalize,noabbrev]{cleveref}

\definecolor{ours}{HTML}{2A78D6}
\definecolor{base}{HTML}{EB6834}
\definecolor{win}{HTML}{1B7F5C}
\newcommand{\sg}{\operatorname{sg}}
\newcommand{\LN}{\operatorname{LN}}
\newcommand{\new}[1]{\textbf{#1}}
\icmltitlerunning{StepJEPA: Step-Level Joint-Embedding Prediction for Latent Reasoning}

\begin{document}

\twocolumn[
  \icmltitle{StepJEPA: Step-Level Joint-Embedding Prediction as a Paradigm\\
    for Latent Reasoning in Language Models}
  \icmlsetsymbol{equal}{*}
  \begin{icmlauthorlist}
    \icmlauthor{Ahmed Elshabrawy}{mbz,nict}
    \icmlauthor{Masao Utiyama}{nict}
    \icmlauthor{Alham Fikri Aji}{mbz}
  \end{icmlauthorlist}
  \icmlaffiliation{mbz}{MBZUAI}
  \icmlaffiliation{nict}{NICT, Japan}
  \icmlcorrespondingauthor{Ahmed Elshabrawy}{ahmed.elshabrawy@mbzuai.ac.ae}
  \icmlkeywords{latent reasoning, JEPA, chain of thought, autoregressive language models, out-of-distribution generalization}
  \vskip 0.3in
]
\printAffiliationsAndNotice{}

\begin{abstract}
Chain-of-thought (CoT) improves reasoning by externalizing intermediate work into text, yet ties computation to noisy token-by-token generation. Existing latent-reasoning paradigms replace written tokens via indirect language-modeling supervision, single-anchor distillation, recurrent-depth looping, token-space step decoding, or two-stage compression, yet lack a single-stage representation-space objective \textit{directly} supervising step-to-step logical transitions. We propose \textbf{StepJEPA}, which encodes logical steps into compact latent blocks and trains one shared autoregressive transformer with a Joint-Embedding Predictive Architecture (JEPA) transition loss, collapse regularization, joint step decoding, and adaptive halting. Forecasting the next logical state directly in representation space reduces token-selection noise, adapts depth to difficulty, and learns transferable dynamics. Across 19M--164M strictly causal autoregressive models, StepJEPA solves deductive reasoning (ProntoQA) from random initialization where prior baselines remain at chance, extrapolates beyond training depth where matched No-JEPA and fixed-budget latent methods collapse, and stays competitive on pretrained 124M GPT-2 GSM8k while pulling far ahead of fixed-budget latent baselines on out-of-distribution depth extrapolation and multi-hop graph and DAG reasoning.
\end{abstract}

\section{Introduction}
\label{sec:intro}
\begin{figure*}[t]
\centering
\includegraphics[width=0.88\textwidth]{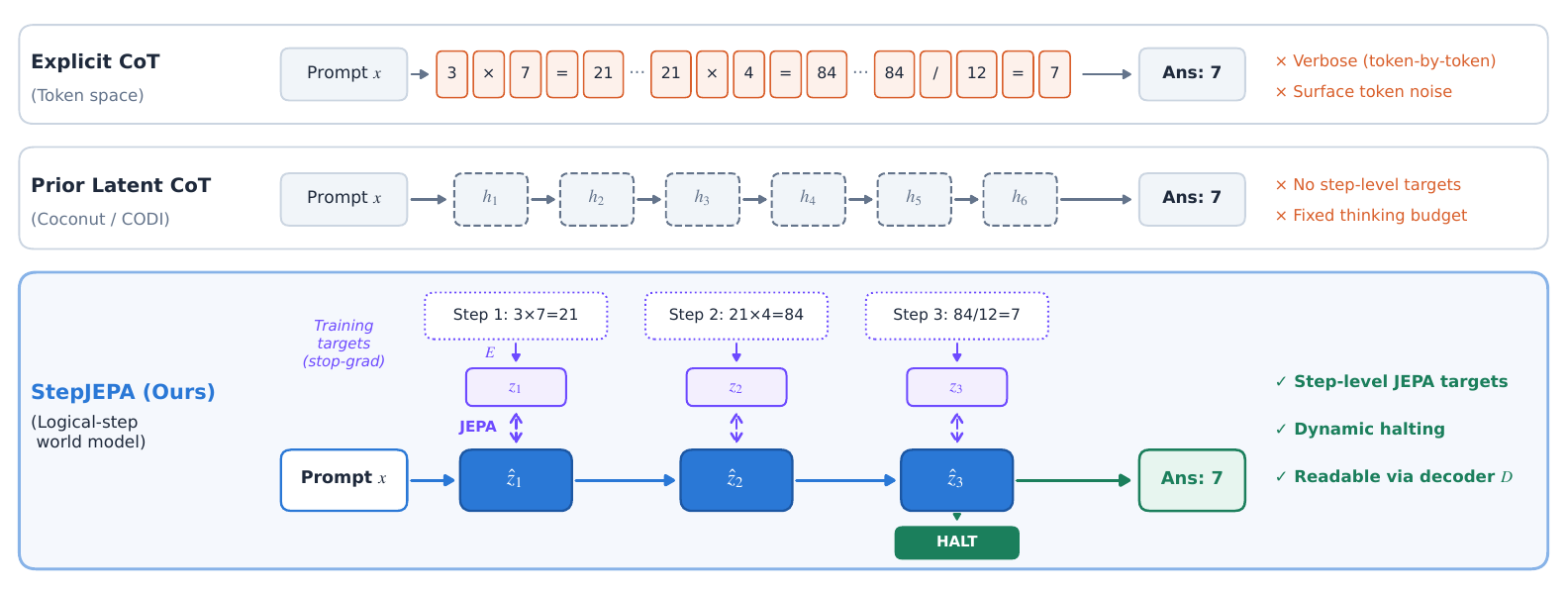}
\vspace{-1.5mm}
\caption{\textbf{Overview of StepJEPA in contrast to prior reasoning paradigms.} \textbf{Top:} Explicit chain-of-thought serializes reasoning one noisy token at a time. \textbf{Middle:} Widely used latent-CoT baselines (e.g., Coconut, CODI, PCCoT) run a fixed budget of continuous states without explicit representation-space step transitions or dynamic halting. \textbf{Bottom:} \textbf{StepJEPA} compresses each logical step into a compact target encoding $z_t$ during training and trains a shared causal transformer to predict the next logical step $\hat{z}_t$ directly in latent space via JEPA, halting dynamically once the conclusion is reached while remaining readable through a step decoder $D$.}
\label{fig:hook}
\end{figure*}

Explicit chain-of-thought (CoT) reasoning \citep{wei2022cot} unlocked multi-step problem solving by allocating test-time compute through generated text, yet reasoning in surface token space (\cref{fig:hook}, top) is both \emph{sequentially inefficient} (forward passes scale with verbal verbosity rather than logical depth) and exposed to \emph{token-level nuisance noise} that triggers irreversible compounding errors. Recent work therefore moves intermediate reasoning into continuous latent space (\cref{fig:hook}, middle) via indirect language-modeling supervision (Coconut; \citealp{hao2024coconut}), fixed-budget single-anchor distillation (CODI, PCCoT; \citealp{shen2025codi,wu2025pccot}), recurrent-depth looping \citep{geiping2025huginn,zhu2025ouro}, chunk compression \citep{tan2025colar}, text-reconstructed steps \citep{wei2025simcot,wang2026plat}, or two-stage codecs \citep{kang2025ladir}. Across these paradigms, a core gap remains: \emph{no single-stage objective explicitly supervises how one logical thought transitions to the next in representation space with collapse prevention and adaptive depth} (\cref{sec:related}).

We posit that the natural unit of latent thought is neither a single token nor an unconstrained hidden vector, but one \emph{logical reasoning step}, such as expanding a search frontier or executing an arithmetic derivation. Among representation-learning paradigms \citep{kingma2014vae,chen2020simclr,he2022mae,ha2018worldmodels,hafner2020dreamer}, Joint-Embedding Predictive Architectures (JEPA; \citealp{lecun2022path,assran2023ijepa,bardes2024vjepa,balestriero2025lejepa}) stand out for learning world-model dynamics by predicting future states directly in abstract feature space rather than reconstructing noisy surface details. 

Drawing on this parallel, we propose \textbf{StepJEPA} (\cref{fig:hook}, bottom), which compresses each logical step into a compact latent block and trains a single unified strictly causal autoregressive transformer (serving as step encoder, latent predictor, step decoder, and answer readout via structured masks and role embeddings) to forecast the next logical state directly in latent space. This formulation improves compact-model sample efficiency, enables dynamic adaptive halting, keeps long out-of-distribution (OOD) rollouts on-manifold, and does not compete with or completely replace prior techniques but rather augments and strengthens them, composing cleanly with auxiliary CoT supervision and parallel Jacobi updates. In contrast with most previous latent reasoning approaches, it comes with the added advantage that StepJEPA latent states decode into clean interpretable steps that allow us to inspect the reasoning chain and pinpoint exact failures in reasoning. Across 19M--164M autoregressive models and a broad suite of logical, planning, commonsense, and math benchmarks, our experiments establish four core findings:
\begin{itemize}
  \vspace{-1.5mm}
  \itemsep0em
  \item \new{Compact models learn generalizable rules from scratch.} Without pretraining, a 19M StepJEPA model solves unseen deductive logic (99.25\% on ProntoQA) where explicit CoT and prior latent baselines stay at chance ($\approx$50\%), and a 164M model generalizes to mazes $1.67\times$ longer than training paths (74.2\%) where an identical model without JEPA collapses to 0.0\%.
  \item \new{The JEPA loss stabilizes multi-step thinking while keeping thoughts readable.} Removing only $\mathcal{L}_{\text{JEPA}}$ under the exact same backbone, loop, and step decoder sharply degrades from-scratch logic, pretrained math, and deep unseen chains (more than halving accuracy on deeper arithmetic graphs, 60.2\% $\to$ 27.0\%), while the shared decoder translates imagined latents into readable text on demand.
  \item \new{Strong real-world math at lower latency, with decisive gains on deeper problems.} On grade-school math (124M GSM8k), adaptive StepJEPA reaches 44.96\% (ahead of explicit CoT, CODI, and Coconut) at nearly $3\times$ lower latency than text CoT and composes with Jacobi decoding. While PCCoT leads short GSM8k problems (48.60\%), StepJEPA more than doubles PCCoT on deeper unseen arithmetic graphs (60.2\% vs.\ 23.8\%) and beats both CODI and PCCoT on multi-hop graph planning (94.4\% on ProsQA, matching our No-JEPA control).
  \item \new{Verified components and smooth scaling with model size.} Ablations confirm that $\mathcal{L}_{\text{JEPA}}$, multi-step self-rollouts, and Sketched Isotropic Gaussian Regularization (SIGReg) drive the core gains rather than auxiliary CoT or decoder gradients. Across 19M--124M causal models, StepJEPA scales smoothly across math, zero-shot transfer, and deeper unseen reasoning, with a 19M model already outperforming 48M CODI and PCCoT on GSM8k.
\end{itemize}

Viewed more broadly, StepJEPA reframes latent reasoning as \emph{explicit step-level transition prediction in representation space}: a language model learns a recurrent transition predictor over logical states, thinking step by step without verbalizing until a final answer is needed. Decoupling the unit of thought from the unit of text lets internal compute scale with logical depth rather than surface verbosity, suggesting the exciting possibility that step-level joint-embedding prediction may unlock structured multi-step reasoning in far smaller models than is the current norm.

\section{Related Work}
\label{sec:related}
\textbf{Latent, step-supervised, and recurrent reasoning.} Prior methods internalize reasoning via vertical or stepwise CoT removal (Implicit CoT, iCoT; \citealp{deng2023icotkd,deng2024icot}), discrete compute tokens (pause/filler tokens; \citealp{goyal2023pause,pfau2024filler}), or fixed-budget continuous states supervised by downstream text or single-anchor distillation (Coconut \citealp{hao2024coconut}, CODI \citealp{shen2025codi}, PCCoT \citealp{wu2025pccot}). Newer variants attach step decoders (SIM-CoT, PLaT; \citealp{wei2025simcot,wang2026plat}), two-stage VAE/diffusion codecs (LaDiR; \citealp{kang2025ladir}), dynamic chunk compression (CoLaR; \citealp{tan2025colar}), looped layers (Huginn, Ouro; \citealp{geiping2025huginn,saunshi2025looped,zhu2025ouro}), tiny grid solvers (HRM, TRM; \citealp{wang2025hrm,jolicoeur2025trm}), or sentence-concept models (LCM; \citealp{barrault2024lcm}). StepJEPA instead unifies single-stage stop-gradient JEPA transitions, SIGReg, joint step decoding, and adaptive halting inside one causal backbone (\cref{app:method}).

\textbf{JEPA and world models for language.} JEPA predicts abstract representations rather than raw inputs \citep{lecun2022path,assran2023ijepa,bardes2024vjepa}, with LeJEPA \citep{balestriero2025lejepa} and LeWorldModel \citep{maes2026lewm} replacing EMA via SIGReg. Prior language-JEPA works either retain explicit token-level inference (LLM-JEPA \citealp{llmjepa2025}, STP \citealp{huang2026stp}, DLLM-JEPA \citealp{dllmjepa2026}, NextLat \citealp{teoh2025nextlat}) or rely on separate or frozen modules without logical-step halting (VL-JEPA \citealp{chen2025vljepa}, JEPA-Reasoner \citealp{jepareasoner2025}, JELAR \citealp{wu2026jelar}). Closest in high-level motivation is Semantic Step Prediction (SSP; \citealp{yuan2026ssp}), which applies STP's cosine-displacement regularizer at step delimiters during standard token-level fine-tuning and probes frozen teacher-forced trajectories post hoc via constant-velocity linear or MLP extrapolation. Crucially, as its author notes, SSP implements neither a target encoder, a learned JEPA transition predictor, nor closed-loop latent inference; it yields no accuracy gain over vanilla fine-tuning, and leaves adaptive depth, OOD extrapolation, and comparisons against latent-reasoning baselines unaddressed. 

To our knowledge, StepJEPA is the first paradigm to elevate logical-step prediction into a closed-loop, adaptive-depth, single-backbone causal JEPA-based reasoning mechanism with SIGReg collapse control and readable step decoding, demonstrating decisive OOD superiority over fixed-budget latent baselines (while matching or surpassing explicit CoT at 124M OOD and solving from-scratch ProntoQA where CoT remains at chance) alongside the causal necessity of the JEPA loss, $\mathcal{L}_{\text{JEPA}}$ (\cref{tab:related}, \cref{app:method}).

\section{Method: StepJEPA}
\label{sec:method}
Each training example consists of a problem $x$, an ordered sequence of logical reasoning steps $s_0, s_1, \dots, s_T$ (where $s_0$ represents the initial state or a neutral start marker), and a final answer $y$. Our formulation directly operationalizes the central hypothesis: compress each logical step $s_t$ into a compact latent block $z_t \in \mathbb{R}^{K \times d}$ consisting of $K$ continuous vectors of hidden dimension $d$, supervise step-to-step transitions in latent space via JEPA, and unify all roles inside a single autoregressive transformer (\cref{fig:masks}).

\subsection{Unified Single-Backbone Architecture}
\label{sec:arch}
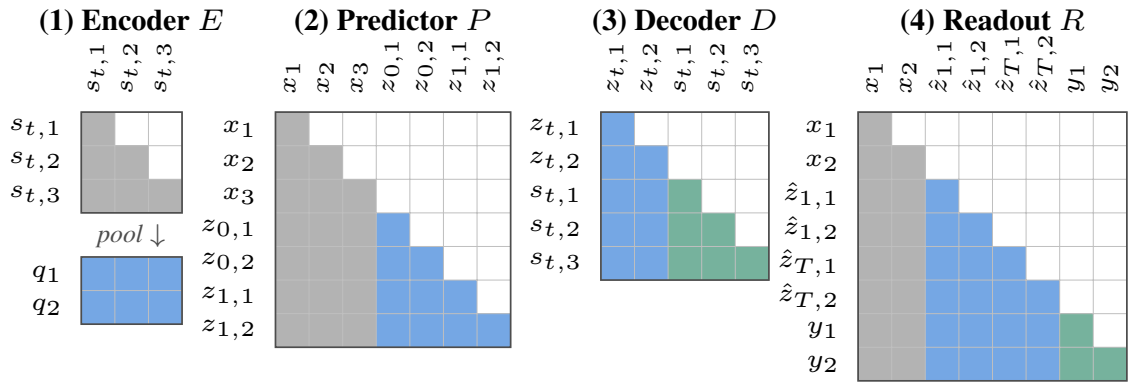
\begin{figure}[t]
\centering
\resizebox{0.92\columnwidth}{!}{\begin{tikzpicture}[scale=0.86]

\begin{scope}[xshift=0.0cm]
\node[font=\scriptsize\bfseries] at (0.48,0.86) {(1) Encoder $E$};
\foreach \i/\lbl in {0/$s_{t,1}$, 1/$s_{t,2}$, 2/$s_{t,3}$} {
  \node[font=\tiny, anchor=east] at (-0.03, -\i*0.32 - 0.16) {\lbl};
  \node[font=\tiny, rotate=90, anchor=west] at (\i*0.32 + 0.16, 0.03) {\lbl};
}
\foreach \r in {0,1,2} {
  \foreach \c in {0,...,\r} {
    \fill[black!30] (\c*0.32, -\r*0.32) rectangle ++(0.32,-0.32);
  }
}
\draw[step=0.32, black!25, very thin] (0,0) grid (0.96,-0.96);
\draw[black!70] (0,0) rectangle (0.96,-0.96);
\node[font=\tiny\itshape, black!70] at (0.48,-1.18) {pool $\downarrow$};
\foreach \i/\lbl in {0/$q_1$, 1/$q_2$} {
  \node[font=\tiny, anchor=east] at (-0.03, -1.38 - \i*0.32 - 0.16) {\lbl};
  \foreach \c in {0,1,2} {
    \fill[ours!65] (\c*0.32, -1.38 - \i*0.32) rectangle ++(0.32,-0.32);
  }
}
\draw[step=0.32, black!25, very thin, yshift=-1.38cm] (0,0) grid (0.96,-0.64);
\draw[black!70] (0,-1.38) rectangle (0.96,-2.02);
\end{scope}

\begin{scope}[xshift=1.85cm]
\node[font=\scriptsize\bfseries] at (1.12,0.86) {(2) Predictor $P$};
\foreach \i/\lbl in {0/$x_1$, 1/$x_2$, 2/$x_3$, 3/$z_{0,1}$, 4/$z_{0,2}$, 5/$z_{1,1}$, 6/$z_{1,2}$} {
  \node[font=\tiny, anchor=east] at (-0.03, -\i*0.32 - 0.16) {\lbl};
  \node[font=\tiny, rotate=90, anchor=west] at (\i*0.32 + 0.16, 0.03) {\lbl};
}
\foreach \r in {0,1,2} {
  \foreach \c in {0,...,\r} {
    \fill[black!30] (\c*0.32, -\r*0.32) rectangle ++(0.32,-0.32);
  }
}
\foreach \r in {3,4,5,6} {
  \foreach \c in {0,1,2} {\fill[black!30] (\c*0.32, -\r*0.32) rectangle ++(0.32,-0.32);}
  \foreach \c in {3,...,\r} {\fill[ours!65] (\c*0.32, -\r*0.32) rectangle ++(0.32,-0.32);}
}
\draw[step=0.32, black!25, very thin] (0,0) grid (2.24,-2.24);
\draw[black!70] (0,0) rectangle (2.24,-2.24);
\end{scope}

\begin{scope}[xshift=4.95cm]
\node[font=\scriptsize\bfseries] at (0.80,0.86) {(3) Decoder $D$};
\foreach \i/\lbl in {0/$z_{t,1}$, 1/$z_{t,2}$, 2/$s_{t,1}$, 3/$s_{t,2}$, 4/$s_{t,3}$} {
  \node[font=\tiny, anchor=east] at (-0.03, -\i*0.32 - 0.16) {\lbl};
  \node[font=\tiny, rotate=90, anchor=west] at (\i*0.32 + 0.16, 0.03) {\lbl};
}
\foreach \r in {0,1} {
  \foreach \c in {0,...,\r} {\fill[ours!65] (\c*0.32, -\r*0.32) rectangle ++(0.32,-0.32);}
}
\foreach \r in {2,3,4} {
  \foreach \c in {0,1} {\fill[ours!65] (\c*0.32, -\r*0.32) rectangle ++(0.32,-0.32);}
  \foreach \c in {2,...,\r} {\fill[win!60] (\c*0.32, -\r*0.32) rectangle ++(0.32,-0.32);}
}
\draw[step=0.32, black!25, very thin] (0,0) grid (1.60,-1.60);
\draw[black!70] (0,0) rectangle (1.60,-1.60);
\end{scope}

\begin{scope}[xshift=7.40cm]
\node[font=\scriptsize\bfseries] at (1.28,0.86) {(4) Readout $R$};
\foreach \i/\lbl in {0/$x_1$, 1/$x_2$, 2/$\hat z_{1,1}$, 3/$\hat z_{1,2}$, 4/$\hat z_{T,1}$, 5/$\hat z_{T,2}$, 6/$y_1$, 7/$y_2$} {
  \node[font=\tiny, anchor=east] at (-0.03, -\i*0.32 - 0.16) {\lbl};
  \node[font=\tiny, rotate=90, anchor=west] at (\i*0.32 + 0.16, 0.03) {\lbl};
}
\foreach \r in {0,1} {
  \foreach \c in {0,...,\r} {\fill[black!30] (\c*0.32, -\r*0.32) rectangle ++(0.32,-0.32);}
}
\foreach \r in {2,3,4,5} {
  \foreach \c in {0,1} {\fill[black!30] (\c*0.32, -\r*0.32) rectangle ++(0.32,-0.32);}
  \foreach \c in {2,...,\r} {\fill[ours!65] (\c*0.32, -\r*0.32) rectangle ++(0.32,-0.32);}
}
\foreach \r in {6,7} {
  \foreach \c in {0,1} {\fill[black!30] (\c*0.32, -\r*0.32) rectangle ++(0.32,-0.32);}
  \foreach \c in {2,3,4,5} {\fill[ours!65] (\c*0.32, -\r*0.32) rectangle ++(0.32,-0.32);}
  \foreach \c in {6,...,\r} {\fill[win!60] (\c*0.32, -\r*0.32) rectangle ++(0.32,-0.32);}
}
\draw[step=0.32, black!25, very thin] (0,0) grid (2.56,-2.56);
\draw[black!70] (0,0) rectangle (2.56,-2.56);
\end{scope}
\end{tikzpicture}}
\vspace{-1.5mm}
\caption{Strictly causal structured attention masks enabling one autoregressive transformer to perform all four roles ($K=2$ slots/step): \textbf{(1)} causal encoder $E$ with post-transformer pooling queries $q_{1:K}$, \textbf{(2)} causal block predictor $P$, \textbf{(3)} step decoder $D$, and \textbf{(4)} causal readout $R$ over $[x; \hat{z}_{1:T}]$.}
\label{fig:masks}
\end{figure}

Rather than maintaining separate networks, a single autoregressive transformer $f_\theta$ of hidden width $d$ performs four distinct roles separated solely by learned input-type (role) embeddings (five core role embeddings $e_{\text{prob}}, e_{\text{step}}, e_{\text{lat}}, e_{\text{canv}}, e_{\text{ans}} \in \mathbb{R}^d$ plus $e_{\text{draft}} \in \mathbb{R}^d$ for inner refinement) added to input vectors and by structured attention masks intersected with the backbone's causal mask (building on mask-unified architectures; \citealp{dong2019unilm,zhou2024transfusion,sun2024latentlm}). Because the backbone remains a standard causal transformer, StepJEPA applies directly to off-the-shelf pretrained models (e.g., GPT-2 via LoRA; \citealp{hu2022lora}) with no architectural changes:
\begin{enumerate}
  \vspace{-1.5mm}
  \itemsep0em
  \item \textbf{Step Encoder $E(s_t) \to z_t$.} Adds $e_{\text{step}}$ to the token embeddings of logical step $s_t$ (with dropout disabled during target encoding so JEPA targets are noise-free), passes them through $f_\theta$, and applies $K$ learned attention-pooling queries followed by parameter-free LayerNorm $\LN(\cdot)$ (exact pooling formula in \cref{app:method}). An entire 15--40 token logical step is thus condensed into a canonicalized block $z_t \in \mathbb{R}^{K \times d}$ ($K \in \{2, 3\}$) lying on the zero-mean, unit-variance sphere. The initial state $s_0$ encodes the starting configuration on state-tracking tasks (e.g., starting maze cell) or a neutral marker on math and QA.
  \item \textbf{Latent Predictor $P(x, z_{0:t}) \to (\hat{z}_{t+1}, p^{\text{done}}_{t+1})$.} Projects each prior latent block into the transformer stream via a linear input map $W_{\text{in}} z_j + e_{\text{lat}}$ ($W_{\text{in}} \in \mathbb{R}^{d \times d}$), concatenates the problem tokens $x$ (added with $e_{\text{prob}}$) and prior latent blocks $z_0, \dots, z_t$, and runs $f_\theta$ under a \emph{block-causal} mask: vectors within step block $t$ attend to the problem, all preceding blocks, and prior slots within block $t$, but never to future blocks (\cref{fig:masks}). A linear output projection $W_{\text{out}} \in \mathbb{R}^{d \times d}$ maps the output states $h_t \in \mathbb{R}^{K \times d}$ of block $t$ back onto the canonical latent sphere, $\hat{z}_{t+1} = \LN(W_{\text{out}} h_t)$. In parallel, a linear halting head takes the slot-mean hidden state $\bar{h}_t = \frac{1}{K}\sum_{k=1}^K h_{t,k}$ and outputs a stopping probability $p^{\text{done}}_{t+1} = \sigma(w_h^\top \bar{h}_t + b_h) \in [0, 1]$.
  \item \textbf{Step Decoder $D(z_t) \to \hat{s}_t$.} Prepends the projected latent block $W_{\text{in}} z_t + e_{\text{lat}}$ before shifted step tokens (tagged with $e_{\text{canv}}$) and autoregressively reconstructs the text of step $s_t$ via standard teacher-forced cross-entropy. We train $D$ jointly by default so any imagined thought $\hat{z}_t$ remains directly readable in plain text at test time (97\%--100\% exact-match reconstruction from gold encodings $z_t$ across tasks); blocking $D$'s gradients as a detached probe or removing $D$ entirely preserves accuracy (\cref{sec:jepa_mech}, \cref{tab:decodability}), confirming that $\mathcal{L}_{\text{JEPA}}$ rather than decoder pressure drives reasoning.
  \item \textbf{Answer Readout $R(x, \hat{z}_{1:T}) \to \hat{y}$.} Autoregressively emits only the concise final answer $y$ conditioned on the problem $x$ and the imagined latent chain (reading from $[x; \hat{z}_{1:T}]$ on short-answer benchmarks such as GSM8k, ProsQA, StrategyQA, and ProntoQA, or emitting one coordinate pair per step from $\hat{z}_t$ for long structured outputs such as full maze trajectories; \cref{fig:masks}).
\end{enumerate}

\textbf{Parameter overhead.} Sharing one backbone across all four roles adds only lightweight projections ($W_{\text{in}}, W_{\text{out}}$), role embeddings, pooling queries, draft vectors, and the halting head: $2d^2 + O(d)$ parameters independent of depth ($<1\%$ of total parameters across 19M--124M backbones; \cref{app:setup}).

\subsection{Training Objective}
\label{sec:obj}
Given encoded ground-truth steps $z_{0:T} = \{E(s_t)\}_{t=0}^T$, the predictor forecasts next-step latents $\hat{z}_{1:T}$ and halting probabilities $p^{\text{done}}_{1:T}$. Our objective combines six loss groups (unit default weights; rollout summation in \cref{app:method}, ablations in \cref{tab:app_v3}):
\begin{align}
  \mathcal{L}_{\text{StepJEPA}} &= \lambda_J \mathcal{L}_{\text{JEPA}}
    + w_{\text{dec}}(\mathcal{L}_{\text{rec}} + \mathcal{L}_{\text{pdec}})
    + \mathcal{L}_{\text{done}} + \mathcal{L}_{\text{ans}} \nonumber\\
    &\quad + \mathcal{L}_{\text{SIGReg}}(z) + \mathcal{L}_{\text{SIGReg}}(\hat{z})
    + w_{\text{cot}}\mathcal{L}_{\text{cot}}.
\label{eq:loss}
\end{align}

\textbf{Step-to-step JEPA transition loss ($\mathcal{L}_{\text{JEPA}}$).} Because both encoder outputs $z_t$ and predictor outputs $\hat{z}_t$ are already canonicalized by parameter-free LayerNorm $\LN(\cdot)$, predicted latents are trained directly to match stop-gradient target encodings in normalized representation space:
\begin{equation}
  \mathcal{L}_{\text{JEPA}} = \frac{1}{T K d} \sum_{t=1}^{T} \bigl\| \hat{z}_t - \sg(z_t) \bigr\|_2^2,
\end{equation}
where $\sg(\cdot)$ stops target-encoding gradients so the encoder cannot collapse toward the predictor (context inputs $z_{0:T-1}$ to $P$ and gold latents to $R$ retain gradients so $E$ remains semantically grounded even without a step decoder).

\textbf{Isotropy and collapse prevention ($\mathcal{L}_{\text{SIGReg}}$).} Following LeJEPA \citep{balestriero2025lejepa}, we prevent representational collapse without an EMA target network via Sketched Isotropic Gaussian Regularization (SIGReg), which projects latent slot vectors onto random unit directions and matches their empirical characteristic functions to $\mathcal{N}(0, 1)$ (exact Epps--Pulley quadrature form in \cref{app:method}). In high dimension (e.g., $d = 768$), isotropic Gaussian vectors concentrate sharply on the zero-mean, unit-variance sphere of radius $\sqrt{d}$, making parameter-free LayerNorm and SIGReg geometrically compatible: LayerNorm fixes per-vector scale while SIGReg pushes all directions across the batch toward full-rank isotropy. Applying SIGReg (pooled across all valid batch, step, and slot vectors) to \emph{both} encoded targets $z$ and predicted thoughts $\hat{z}$ prevents rollout contraction across multi-step chains (\cref{app:v3}, \cref{tab:app_v3}).

\textbf{Functional step decodability ($\mathcal{L}_{\text{rec}} + \mathcal{L}_{\text{pdec}}$).} Standard teacher-forced autoregressive cross-entropy trains the shared step decoder $D$ to reconstruct step text $s_t$ from gold encodings ($\mathcal{L}_{\text{rec}} = \frac{1}{T+1}\sum_{t=0}^T \operatorname{CE}(D(z_t), s_t)$) and predicted latents ($\mathcal{L}_{\text{pdec}} = \frac{1}{T}\sum_{t=1}^T \operatorname{CE}(D(\hat{z}_t), s_t)$), with default $w_{\text{dec}} = 1$.

\textbf{Dynamic halting without length ceilings ($\mathcal{L}_{\text{done}}$).} Drawing on adaptive computation time \citep{graves2016act,dehghani2019ut,banino2021pondernet}, binary cross-entropy trains the halting probability $p^{\text{done}}_t$ to equal $1$ on the final logical step ($t = T$) and $0$ on all earlier steps ($t < T$; exact formula in \cref{app:method}). To prevent the model from memorizing a fixed maximum training depth, we omit absolute step-index embeddings \citep{kazemnejad2023nope} and, following shifted and randomized positional encodings \citep{kiyono2021shape,ruoss2023randomized}, apply \emph{positional jitter}, adding a per-example random integer offset $\delta \sim \operatorname{Uniform}\{0, \dots, J\}$ to the starting positional index of all latent blocks during training so attention relies on relative step-to-step progress rather than absolute sequence positions.

\textbf{Closing exposure bias in rollout and readout ($\mathcal{L}_{\text{ans}}$).} To prevent compounding errors when the predictor conditions on its own outputs at test time, we run $S_{\text{ro}}$-pass multi-step self-rollouts ($S_{\text{ro}} \in \{2, 3\}$): after the teacher-forced pass $\hat{z}^{(1)}_{1:T} = P(x, z_{0:T-1})$, each subsequent pass feeds back predicted latents, $\hat{z}^{(s)}_{1:T} = P(x, [z_0, \hat{z}^{(s-1)}_{1:T-1}])$, propagating up to $S_{\text{ro}}-1$ steps of compounded autonomous error with end-to-end gradients and full step, halting, and answer losses on the final rollout (optionally combined with detached scheduled sampling at rate $p_{\text{ro}}$; \citealp{bengio2015scheduled}). The answer readout is supervised on a weighted mixture of predicted and gold latents, $\mathcal{L}_{\text{ans}} = (1 - \alpha_g)\,\operatorname{CE}(R(x, \hat{z}_{1:T}), y) + \alpha_g\,\operatorname{CE}(R(x, z_{1:T}), y)$ ($\alpha_g$ sensitivity in \cref{tab:app_v3}), keeping readout aligned with clean latents while robust to autonomous rollouts.

\textbf{Composability with auxiliary explicit CoT ($\mathcal{L}_{\text{cot}}$).} To demonstrate orthogonality with standard language-modeling supervision (analogous to the CoT teacher branch in CODI and PCCoT), the shared backbone can optionally co-train on written chain-of-thought traces via standard autoregressive cross-entropy with weight $w_{\text{cot}}$ (exact form in \cref{app:method}). Crucially, this loss is used \emph{only during training}; inference remains completely latent, emitting only the final answer.

\subsection{Inference, Halting Trade-Offs, and Jacobi Decoding}
\label{sec:paradigms}
Because StepJEPA specifies how latent thoughts are supervised rather than a rigid decoding schedule, one causal backbone supports two inference modes (\cref{alg:stepjepa}, \cref{app:method}):

\textbf{Adaptive sequential refinement (default).} Beginning from the initial state encoding $z_0 = E(s_0)$, the predictor generates logical-step latents $\hat{z}_1, \hat{z}_2, \dots$ one step at a time. To give the causal transformer multiple internal computation passes per logical step, each step block $t$ is paired with $K$ private \emph{draft slots} initialized from a learned null draft $d^{(0)} \in \mathbb{R}^{K \times d}$ and refined over $M$ weight-tied passes (default $M = 2$): on pass $m \in \{1, \dots, M\}$, the draft slots attend only to $x$, committed latents $\hat{z}_{0:t-1}$, and themselves, producing $\hat{z}^{(m)}_t = \LN(W_{\text{out}} h^{(m)}_t)$ which is fed back as $W_{\text{in}} \hat{z}^{(m)}_t + e_{\text{draft}}$ for pass $m + 1$ (with gradients flowing through all $M$ passes during training). Because draft slots are private to their own step, all $M$ passes over all steps execute in parallel during training and match sequential step-by-step refinement at inference. After $M$ passes, the halting head evaluates $p^{\text{done}}_t = \sigma(w_h^\top \bar{h}^{(M)}_t + b_h)$; the loop stops as soon as $p^{\text{done}}_t > \tau$ (default $\tau = 0.5$) or when the number of steps exceeds a maximum thinking budget ($t = T_{\max}$), and the answer readout $R$ emits the final answer tokens. Exact per-example sequential forward passes (Number of Function Evaluations, NFE) thus follow $\text{NFE}_{\text{seq}} = 1 + \hat{T} \cdot M + N_{\text{ans}}$ (one pass to encode $s_0$, $\hat{T}$ dynamically chosen steps at $M$ passes each, and $N_{\text{ans}}$ greedy answer-decoding passes until end-of-sequence; e.g., $1 + 3.33 \times 2 + 2.24 = 9.90$ per-example NFE on GSM8k).

\textbf{Halting head vs.\ state-based halting predicate.} Our default halting head reads directly from $\bar{h}^{(M)}_t$ at zero extra forward passes, retaining prefix history needed for non-Markovian word problems (GSM8k, SVAMP, StrategyQA). Because prefix attention can still sense sequence length under extreme $3\times$--$4\times$ depth extrapolation, Markovian state-tracking tasks (e.g., 19M Long Mazes, \cref{tab:gen}) can optionally use a \emph{state-based halting predicate} on $[x; \hat{z}_t]$ alone (masking prior history), trading one extra pass per step for strict length invariance.

\textbf{Parallel deterministic Jacobi decoding.} To verify direct composability with parallel fixed-point and Jacobi latent decoding \citep{bai2019deq,santilli2023jacobi,kou2024cllm,fu2024lookahead,wu2025pccot}, we also evaluate a Jacobi variant on the exact same autoregressive backbone that initializes a fixed horizon of $H$ latent blocks and updates all $H$ thoughts simultaneously over $I$ parallel block-causal fixed-point iterations ($\text{NFE}_{\text{jac}} = 1 + I + N_{\text{ans}}$; exact update equations in \cref{app:method}).

\section{Experimental Setup}
\label{sec:setup}
We evaluate StepJEPA exclusively on unified, strictly causal autoregressive transformers from 19M to 164M parameters; exact dataset sizes, depth splits, and hyperparameters are listed in \cref{app:tasks,app:setup}.

\textbf{1. Compact models and depth-extrapolation suites.} We evaluate LLaMA-style causal transformers (SMDM-AR at 19M, 34M, and 48M parameters; \citealp{nie2025smdm}, trained both from random initialization and from SlimPajama pretraining; \citealp{soboleva2023slimpajama}) alongside a 164M causal transformer trained both from scratch and from OpenWebText pretraining. Beyond ProntoQA deductive syllogisms \citep{saparov2023prontoqa}, we construct four fixed-vocabulary procedural suites that isolate reasoning depth from unseen-token confounds: shortest-path Mazes (including a long-horizon variant exceeding three times training length), arithmetic dependency graphs (iGSM-Small), one-dimensional cellular automata (CA1D), and shortcut-free Multi-Hop Deduction. On each procedural suite, models are tested both within training depth (ID) and on two progressively deeper unseen splits (OOD1 and OOD2).

\textbf{2. Pretrained 124M GPT-2 benchmarks.} To test natural-language reasoning under exact parity with latent-reasoning baselines, we adapt pretrained GPT-2 Small \citep{radford2019gpt2} via LoRA \citep{hu2022lora} matching CODI and PCCoT. We evaluate GSM8k-Aug \citep{deng2024icot,cobbe2021gsm8k} with zero-shot transfer to MultiArith \citep{roy2015multiarith}, SVAMP \citep{patel2021svamp}, and GSM-Hard \citep{gao2023pal}, alongside iGSM-Small, Multi-Hop Deduction, spatial reasoning on StepGame \citep{shi2022stepgame}, commonsense reasoning on StrategyQA-CoT \citep{geva2021strategyqa}, and graph planning on ProsQA \citep{hao2024coconut}.

\textbf{3. Baselines, controls, and protocol.} We compare against Explicit/token CoT, Coconut \citep{hao2024coconut}, CODI \citep{shen2025codi}, PCCoT \citep{wu2025pccot}, Direct Answer (No-CoT), and a matched No-JEPA control disabling only the predictive JEPA loss under an identical backbone, latent loop, step decoder, and readout (plus an encoder-free self-rollout variant in \cref{app:v3}), alongside published GPT-2 scores for SIM-CoT, Coconut, and iCoT. All checkpoints are validation-selected on held-out splits verified to have zero training overlap (\cref{app:setup}).

\section{Validating StepJEPA}
\label{sec:results}
We evaluate StepJEPA along three questions: whether step-level prediction improves depth extrapolation and small-model learning (\cref{sec:ood}); whether the JEPA loss itself drives these gains while keeping thoughts readable (\cref{sec:jepa_mech}); and whether latent structure preserves accuracy and lowers latency on natural-language math (\cref{sec:gsm124}).

\subsection{OOD Extrapolation and Small-Model Efficiency}
\label{sec:ood}
\begin{figure}[t]
\centering
\includegraphics[width=0.98\columnwidth]{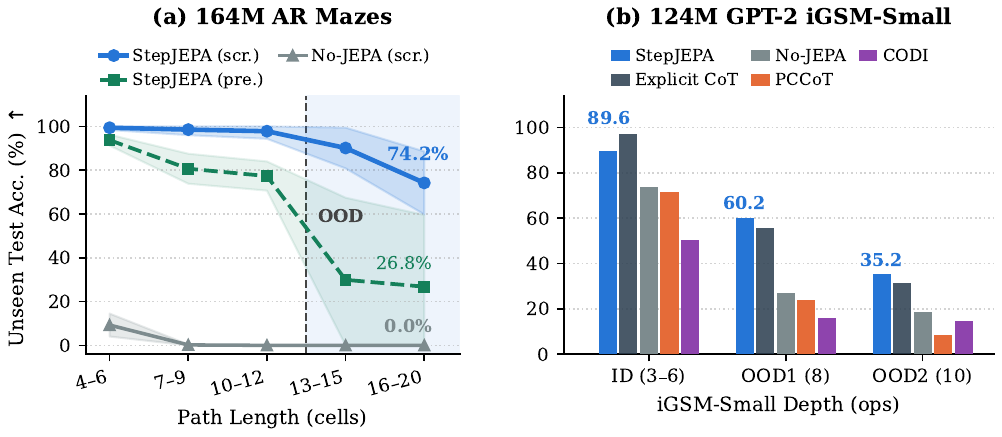}
\vspace{-1.5mm}
\caption{\textbf{Out-of-distribution depth extrapolation on causal autoregressive models.} \textbf{(a)} 164M grid mazes trained on 4--12 cell paths (shaded bands show $\pm$1 standard deviation across three seeds). \textbf{(b)} 124M GPT-2 on iGSM-Small arithmetic graphs trained on 3--6 operations and tested on 8- and 10-operation depths. Error bars are omitted in panel (b) because only Explicit CoT has three completed seeds (plotting its three-seed mean; other methods are single-seed runs).}
\label{fig:main}
\end{figure}

A central prediction of our hypothesis is that supervising explicit step-to-step transitions should generalize beyond the depths seen during training and let even small, unpretrained models learn structured reasoning. Every result in \cref{tab:gen} and \cref{fig:main} is evaluated on held-out test splits with zero training overlap, and every out-of-distribution split strictly exceeds the maximum reasoning depth seen during training.

\begin{table*}[t]
\centering
\caption{Unseen test accuracy (\%) on out-of-distribution extrapolation, commonsense/logic, and sample-efficiency benchmarks across strictly causal autoregressive backbones (19M full-parameter from scratch and 124M GPT-2 LoRA). Best per split is \textbf{bold} (ties go to StepJEPA); second best is \textit{italic}. StepJEPA uses adaptive sequential inference across all rows (parallel Jacobi in \cref{sec:ood}; baseline notes in \cref{app:setup}).}
\label{tab:gen}
\small
\renewcommand{\arraystretch}{1.02}
\resizebox{0.82\textwidth}{!}{
\begin{tabular}{lccccc}
\toprule
Benchmark / Unseen Test Split (Acc.\ \% $\uparrow$) & \textcolor{ours}{\textbf{StepJEPA}} & No-JEPA & CODI & PCCoT & Explicit CoT \\
\midrule
\multicolumn{6}{l}{\textit{Matched 19M AR From Scratch (zero pretraining, full-parameter training)}} \\
\textbf{ProntoQA ($n=800$)} & \textbf{99.25} & \textit{50.75} & 50.38 & 50.13 & 50.00 \\
\textbf{iGSM-Small DAGs (ID / OOD1 / OOD2)} & \textit{96.0} / \textit{26.0} / \textit{16.4} & 78.0 / 17.4 / 12.2 & 15.2 / 8.4 / 7.2 & 22.6 / 9.2 / 9.2 & \textbf{98.2} / \textbf{30.4} / \textbf{21.4} \\
\textbf{1D Cellular Automata (ID / OOD1 / OOD2)} & \textit{94.6} / 0.0 / 0.0 & 83.8 / 0.0 / 0.0 & 92.6 / 0.0 / \textit{0.4} & \textit{94.6} / \textbf{0.6} / \textbf{1.4} & \textbf{100.0} / \textit{0.2} / 0.2 \\
\textbf{Long Mazes (ID 7--12 / OOD 21--40)} & 53.8 / 0.0 & 0.0 / 0.0 & \textbf{73.2} / 0.0 & \textit{70.2} / 0.0 & 53.8 / 0.0 \\
\midrule
\multicolumn{6}{l}{\textit{124M Pretrained GPT-2 (LoRA) Multi-Hop, Commonsense, and OOD Suites}} \\
\textbf{iGSM-Small DAGs (ID / OOD1 / OOD2)} & \textit{89.6} / \textbf{60.2} / \textbf{35.2} & 73.8 / 27.0 / 18.6 & 50.2 / 16.0 / 14.8 & 71.4 / 23.8 / 8.6 & \textbf{97.2} / \textit{55.7} / \textit{31.4} \\
\textbf{StepGame (4 / 8 / 10-hop; audit in \cref{app:neg})} & \textbf{99.4} / \textit{95.6} / \textbf{90.0} & \textit{99.4} / \textbf{96.2} / 55.6 & 97.2 / 81.2 / \textit{65.8} & 98.8 / 43.4 / 33.8 & 88.6 / 86.4 / 57.8 \\
\textbf{Multi-Hop Deduction (ID / OOD1 / OOD2)} & \textit{99.8} / \textit{73.4} / 47.2 & 87.0 / 69.0 / \textbf{64.0} & 51.4 / 50.4 / \textit{47.6} & 51.6 / 54.2 / \textit{47.6} & \textbf{100.0} / \textbf{100.0} / 28.8 \\
\textbf{StrategyQA-CoT ($n=229$)} & \textbf{60.26} & 53.28 & \textit{58.95} & 55.90 & 54.15 \\
\textbf{ProsQA ($n=500$)} & \textit{94.4} & \textbf{95.0} & 79.2 & 87.4 & 85.2 \\
\bottomrule
\end{tabular}
}

\end{table*}

\textbf{From-scratch sample efficiency without pretraining.} Supervising explicit step-to-step transitions lets compact models learn structured reasoning without any language pretraining. When identical 19M causal models are trained from random initialization on ProntoQA syllogisms, StepJEPA solves unseen test proofs at \textbf{99.25\% accuracy} (and finishes faster in wall-clock time than Explicit CoT; \cref{app:setup}), whereas Explicit CoT, CODI, PCCoT, our matched No-JEPA control, and an encoder-free self-rollout No-JEPA control \textbf{all stay at random chance around 50\%}. Tracking per-epoch validation accuracy across all 40 epochs (\cref{app:prontoqa}) rules out over-training: no baseline ever rises above chance, whereas StepJEPA undergoes a sudden generalization (grokking) transition around epoch 28 under the exact same hyperparameters as No-JEPA. 

A similar sample-efficiency advantage appears on from-scratch iGSM-Small arithmetic graphs, where StepJEPA reaches \textbf{96.0\% in-distribution} and retains nontrivial extrapolation on deeper graphs, far ahead of No-JEPA, PCCoT, and CODI and close to Explicit CoT (\cref{tab:gen}). By contrast, extreme-horizon extrapolation remains difficult at this tiny 19M scale: on 1D cellular automata and Long Mazes, models learn in-distribution (with CODI and PCCoT leading in-distribution Long Mazes and StepJEPA matching Explicit CoT while far beating No-JEPA at 0.0\%), yet every 19M from-scratch method collapses on the deepest unseen horizons. Together, these from-scratch trials show that step-level transition prediction substantially lowers the model-size and data threshold for acquiring logical rules, even though very deep rollouts still demand greater capacity (larger sized models).

\textbf{Extrapolating beyond training depth.} Because fixed-budget latent methods pack every derivation into a constant number of hidden vectors, they hit a representational ceiling beyond training depth; StepJEPA avoids this ceiling by unrolling one latent block per step until halting (\cref{fig:main}, \cref{tab:gen}). On 124M GPT-2 iGSM-Small, StepJEPA smoothly scales its predicted steps from 4.4 in-distribution to 8.0 and 9.0 on the two deeper splits, \textbf{more than doubling PCCoT} on 8-operation chains (60.2\% vs.\ 23.8\%) and \textbf{quadrupling it} on 10-operation chains (35.2\% vs.\ 8.6\%) while substantially outperforming CODI and No-JEPA. Remarkably, StepJEPA matches three-seed Explicit CoT within one standard deviation on both deeper splits without emitting a single intermediate token. 

The same separation holds on 10-hop StepGame (90.0\% vs.\ 65.8\% for CODI and 33.8\% for PCCoT; see \cref{app:neg} for a prompt-cue audit) and on 164M spatial mazes trained on 4--12 cell paths (\cref{fig:main}a), where three-seed from-scratch StepJEPA scales its rollout depth monotonically to \textbf{hold 90.1\% on 13--15 cell paths and 74.2\% on 16--20 cell paths} while the matched No-JEPA control collapses to 0.0\%. On Multi-Hop Deduction, StepJEPA similarly scales its rollout depth to reach 99.8\% in-distribution and 73.4\% on 7--8 hop chains (far above chance-level CODI and PCCoT), though its longer rollouts drift near chance on 9--10 hop chains (47.2\%, or 51.8\% with parallel Jacobi), while Explicit CoT drops to 28.8\% because most verbal chains exceed GPT-2's 1,024-token context limit and No-JEPA relies on a 6-step shortcut analyzed below. Overall, allocating compute dynamically per step enables reliable extrapolation beyond training depth, with remaining failures governed by gradual rollout drift rather than a hard slot ceiling.

\textbf{Graph planning, commonsense, and verified causal reliance.} On natural-language graph planning and commonsense benchmarks with pretrained 124M GPT-2 (\cref{tab:gen}), StepJEPA reaches 94.4\% on ProsQA (94.6\% with parallel Jacobi!), beating PCCoT by \textbf{7.0 points}, Explicit CoT by \textbf{9.2 points}, and CODI by \textbf{15.2 points} while matching our LoRA No-JEPA control at 95.0\% (published full-finetuning Coconut and iCoT report 97.0\% and 98.2\%; \citealp{hao2024coconut}). On StrategyQA-CoT, StepJEPA reaches 60.26\% (61.57\% with parallel Jacobi), \textbf{ahead of every matched baseline} on this compact 229-question test set. Because recent mechanistic studies show that unconstrained latent-CoT models can bypass their hidden states via prompt shortcuts \citep{liang2026mechanistic,zhang2025dolatent,qiu2026actuallyreason}, we test causal reliance directly by shuffling or zeroing StepJEPA's imagined latents at test time. Validation accuracy \textbf{immediately collapses to near zero or chance} across every benchmark (from nearly 100\% down to 0\%--2\% on 164M mazes and iGSM-Small, and down to chance on ProntoQA), confirming that the model genuinely reasons through its imagined latent chain.

\subsection{Causal Necessity, Components, and Decoded Thoughts}
\label{sec:jepa_mech}
\textbf{Removing only $\mathcal{L}_{\text{JEPA}}$ collapses autonomous rollouts.} StepJEPA's gains come from the predictive JEPA transition loss itself rather than merely from its unified latent loop or joint step decoder. Our matched No-JEPA control disables only $\mathcal{L}_{\text{JEPA}}$ while keeping the identical backbone, latent loop, step decoder, and readout; to further rule out any train--inference mismatch from gold-latent teacher forcing when $\mathcal{L}_{\text{JEPA}}$ is removed, we also test a strictly encoder-free self-rollout No-JEPA control that encodes only the initial state and unrolls purely autonomously (\cref{tab:jepa_nec} in \cref{app:halting}). Removing $\mathcal{L}_{\text{JEPA}}$ causes steep drops wherever autonomous multi-step dynamics matter: from-scratch 19M ProntoQA falls from 99.25\% to chance (50.75\%, and 50.00\% encoder-free), from-scratch 19M iGSM-Small drops sharply (falling to 39.2\% in-distribution without ground-truth encoder paths), pretrained 19M GSM8k drops by 7.81 points, 164M mazes \textbf{collapse to 0.0\% on unseen depths} both from scratch and from pretraining, and 124M iGSM-Small OOD1 \textbf{plunges from 60.2\% to 27.0\%} despite predicting nearly identical step counts. This isolates transition stability, rather than halting length or decoder capacity, as the mechanism $\mathcal{L}_{\text{JEPA}}$ supplies.

The few exceptions clarify the boundary of this effect. On short, in-distribution ProsQA with a pretrained LoRA backbone, joint step decoding and auxiliary CoT already suffice without $\mathcal{L}_{\text{JEPA}}$ (95.0\% vs.\ 94.4\%). On 124M Multi-Hop Deduction, StepJEPA leads both in-distribution and OOD1 and drops to chance when its latents are shuffled, whereas No-JEPA's higher OOD2 score (64.0\%) reflects halting at the 6-step training ceiling alongside high shuffled accuracy (82.3\%), pointing to a \textbf{prompt shortcut rather than extrapolated dynamics}. Finally, on 164M mazes, training from scratch outperforms OpenWebText pretraining on average because two of three pretrained seeds halt early near the training boundary, yet removing $\mathcal{L}_{\text{JEPA}}$ collapses both regimes to zero.

\textbf{Component ablations and pure-JEPA decodability.} Controlled ablations on pretrained 19M models (\cref{tab:decodability,tab:app_v3} in \cref{app:v3}) isolate which components carry StepJEPA's gains. Adding predicted-latent isotropy regularization, deeper self-rollouts, and jittered step positioning improves GSM8k accuracy by 2.66 points over our base setup. In leave-one-out removals, \textbf{dropping $\mathcal{L}_{\text{JEPA}}$ causes the steepest decline ($-7.81$ points)}, followed by removing deep self-rollouts ($-4.93$ points) and removing all SIGReg regularization ($-3.26$ points), whereas removing the auxiliary explicit-CoT loss costs less than one point ($-0.76$ points). Equally important, \textbf{blocking step-decoder gradients entirely} so the decoder serves purely as a detached interpretability probe closely matches or outperforms reduced-weight and no-decoder variants across 19M to 48M models (\cref{tab:decodability}, retaining auxiliary CoT on GSM8k), confirming that $\mathcal{L}_{\text{JEPA}}$ rather than decoder gradients drives the core latent dynamics.

\textbf{Inspecting imagined thoughts in readable text.} Unlike continuous-CoT baselines lacking a dedicated multi-token step decoder (Coconut, CODI) or approaches relying on separate external decoders \citep{wei2025simcot,wang2026plat,kang2025ladir}, StepJEPA shares its encoder, decoder, and transition predictor inside one causal backbone, allowing any imagined latent $\hat{z}_t$ to be \textbf{decoded directly into plain text without altering inference} (\cref{tab:qualitative} in \cref{app:qualitative}). On GSM8k, decoded latents trace clean step-by-step equations (\texttt{4*20=80} $\to$ \texttt{2*80=160} $\to$ \texttt{80+160+20=260}); on zero-shot MultiArith and SVAMP, halting triggers cleanly after one or two logical steps; and when an error occurs, decoding the imagined sequence pinpoints the exact mistaken step, opening a path to future approaches with targeted precise step-level correction.

\subsection{Pretrained 124M GPT-2: GSM8k and Adaptive Compute}
\label{sec:gsm124}
Having established StepJEPA's gains on algorithmic, logical, and planning benchmarks, we next ask whether explicit step-level JEPA structure hurts accuracy on noisy, natural-language arithmetic. \Cref{tab:gsm124} shows that it does not. On the unseen GSM8k test set with the identical 124M GPT-2 LoRA backbone, adaptive sequential StepJEPA reaches \textbf{44.96\% accuracy while emitting only 2.2 tokens} per problem in 9.90 sequential forward passes. Under the same adapter budget, this outperforms both our reproduction of CODI (42.53\%) and its published score (43.7\%), surpasses Explicit CoT (43.06\%) by \textbf{1.90} points, exceeds Coconut (34.5\%) by \textbf{over 10} points, and slightly edges published Coconut+SIM-CoT (44.8\%; \citealp{wei2025simcot}) while adding adaptive halting. Fixed-budget PCCoT (48.60\%) remains stronger on strictly in-distribution GSM8k, whereas StepJEPA pulls decisively ahead whenever unseen problems exceed training depth (\cref{sec:ood}).

\begin{table*}[t]
\centering
\caption{Full GSM8k unseen test accuracy ($n=1,319$), per-example sequential forward passes (NFE, batch size 1), emitted BPE tokens, and same-GPU batch-1 latency (fp32, merged LoRA on A100-40GB, $n=300$) on 124M GPT-2. StepJEPA runs uncached (conservative upper bound). Among multi-step methods, best per column is \textbf{bold} (ties go to StepJEPA) and second best is \textit{italic}; $\ddagger$published score \citep{wei2025simcot}; bf16 timings and accounting details in \cref{app:setup}.}
\label{tab:gsm124}
\small
\renewcommand{\arraystretch}{1.02}
\resizebox{0.8\textwidth}{!}{
\begin{tabular}{llcccc}
\toprule
Method (124M GPT-2) & Inference Mode & Acc.\ (\%) $\uparrow$ & NFE $\downarrow$ & Emitted $\downarrow$ & A100 B=1 (ms) $\downarrow$ \\
\midrule
Direct Answer (No-CoT) & Direct readout & 17.74 & 6.2 & 6.2 & -- \\
Coconut (3-seed repro) & Indirect latent & 34.5 & 9.2 & $\sim$3 & -- \\
Explicit CoT (repro) & Full text CoT (KV-cached) & 43.06 & 31.3 & 31.3 & 195.2 \\
CODI (official repro / paper) & Fixed 6 latents (KV-cached) & 42.53 / 43.7 & 12.3 & 6.3 & 79.0 \\
SIM-CoT (on Coconut, pub.$\ddagger$) & Aux.\ step dec.\ (fixed) & 44.8 & -- & -- & -- \\
PCCoT (official repro) & Fixed 24 Jacobi (KV-cached) & \textbf{48.60} & \textbf{6.20} & \textit{2.2} & \textbf{39.7} \\
\midrule
\textcolor{ours}{\textbf{StepJEPA (LoRA)}} & Parallel Jacobi (uncached) & 41.62 & \textit{7.22} & \textbf{2.2} & \textit{50.8} \\
\textcolor{ours}{\textbf{StepJEPA (LoRA)}} & Adaptive seq.\ (uncached) & \textit{44.96} & 9.90 & \textbf{2.2} & 66.1 \\
\bottomrule
\end{tabular}
}
\end{table*}

\textbf{Sequential compute, wall-clock latency, and composability.} Because StepJEPA reasons in compact latent blocks rather than emitting verbal chains, it cuts sequential forward passes by \textbf{more than threefold} relative to Explicit CoT (9.90 vs.\ 31.3 passes) and by \textbf{nearly 20\% }relative to CODI (12.3 passes, which includes a short answer prefix alongside its six fixed latents). On a single A100 GPU at batch size 1 (\cref{tab:gsm124}; identical ranking in half-precision bf16, \cref{app:setup}), adaptive StepJEPA finishes in 66.1\,ms even without KV caching, running \textbf{nearly $3\times$ faster than KV-cached Explicit CoT} (195.2\,ms) and $1.2\times$ faster than KV-cached CODI (79.0\,ms). Composing StepJEPA with parallel Jacobi decoding lowers uncached latency further to 50.8\,ms (7.22 forward passes, \textbf{nearly $4\times$ faster than CoT}) at a modest 3.3-point accuracy trade-off. Because 124M GSM8k training is compute-intensive (\cref{tab:app_wallclock}), this run uses our base StepJEPA configuration while full loss-removal ablations are performed at 19M.

\textbf{Zero-shot arithmetic transfer and adaptive halting.} Evaluated zero-shot on unseen arithmetic datasets without retraining, the 124M GSM8k model \textbf{automatically scales its thinking budget to problem difficulty}: it spends fewest forward passes on short problems like SVAMP (47.0\% accuracy in 6.64 passes, \textbf{5.6 points above both CODI and Explicit CoT}) and MultiArith (95.0\% in 7.09 passes, ahead of CODI and CoT at 91.1\%), expands to 9.90 passes on GSM8k, and allocates 11.14 passes on GSM-Hard (9.63\%, matching Explicit CoT; parallel Jacobi numbers in \cref{app:composability}). On these short one- to three-step transfer tasks, fixed-budget PCCoT stays slightly ahead because short derivations fit comfortably within its 24-latent budget, whereas deeper out-of-distribution chains exceed that ceiling and reverse the gap in StepJEPA's favor (\cref{tab:gen}).

\section{Scaling Out-of-Distribution Generalization}
\label{sec:scaling}
The clearest advantage of StepJEPA is not single-task compression, but out-of-distribution generalization: learning reusable step-to-step dynamics that transfer to deeper chains, unseen benchmarks, and smaller models. Examining how these capabilities evolve across our causal transformer ladder from 19M to 164M parameters yields three takeaways (full capacity grid in \cref{app:id_scaling}, \cref{tab:scaling}):

\textbf{1. Generalization emerges early and beats larger fixed-budget baselines.} Even at 19M parameters, step-level transition supervision unlocks reasoning that matched Explicit CoT, distillation, and No-JEPA models fail to acquire from scratch (\cref{tab:gen}). On pretrained GSM8k (\cref{tab:scaling}), compact StepJEPA models consistently outperform much larger fixed-budget latent baselines: \textbf{19M StepJEPA (27.22\%) already surpasses both 48M CODI (24.34\%) and 48M PCCoT (26.84\%)}, while 34M StepJEPA beats both 48M baselines on zero-shot MultiArith and SVAMP. Untruncated Explicit CoT retains an external-scratchpad advantage at the smallest scales before StepJEPA overtakes it at 124M on GSM8k and matches three-seed Explicit CoT within one standard deviation on deep iGSM-Small chains.

\begin{figure}[t]
\centering
\includegraphics[width=0.98\columnwidth]{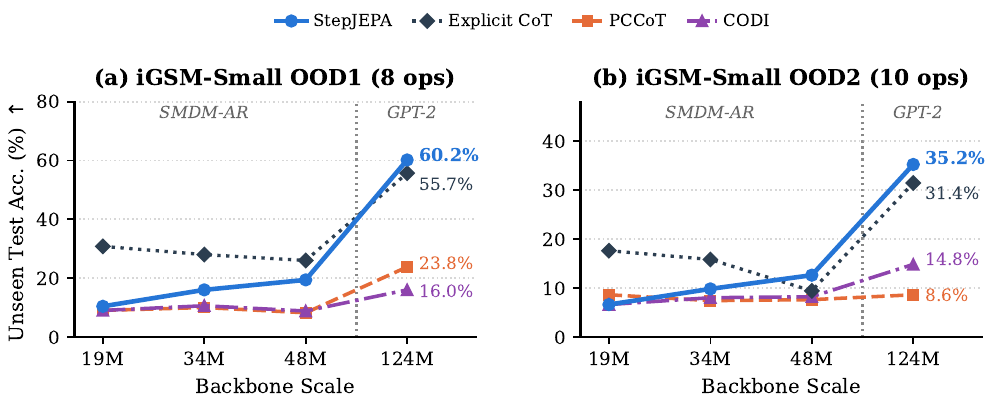}
\vspace{-2.0mm}
\caption{\textbf{OOD depth scaling across 19M--124M autoregressive models} on unseen iGSM-Small \textbf{(a)} OOD1 (8 ops) and \textbf{(b)} OOD2 (10 ops), comparing StepJEPA, Explicit CoT, PCCoT, and CODI (dotted line separates SlimPajama SMDM-AR from OpenWebText GPT-2; \cref{tab:scaling}).}
\label{fig:scaling_ladder}
\end{figure}

\textbf{2. Depth extrapolation and transfer scale smoothly with capacity.} Across our 19M, 34M, 48M, and 124M LoRA scaling ladder (\cref{fig:scaling_ladder}, \cref{tab:scaling}; full-parameter 19M iGSM-Small is reported separately in \cref{tab:gen}), StepJEPA \textbf{improves monotonically} on iGSM-Small in-distribution (34.2\% to 89.6\%), 8-operation OOD1 (10.4\% to 60.2\%), and 10-operation OOD2 (6.6\% to 35.2\%), whereas CODI and PCCoT stall at compact scales and degrade sharply beyond training depth even at 124M. Zero-shot arithmetic transfer climbs just as smoothly across all four sizes, rising from \textbf{63.9\% to 95.0\% on MultiArith} and from 28.8\% to 47.0\% on SVAMP. Even accounting for the backbone family shift between SlimPajama SMDM-AR (19M--48M) and OpenWebText GPT-2 (124M), consistent monotonic gains within each backbone family strongly motivate larger backbones.

\textbf{3. Why larger backbones and multi-task training are the natural next step.} Fixed-budget latent methods face a structural bottleneck: compressing an arbitrarily deep derivation into a constant number of latent vectors forces information loss beyond training depth regardless of model size. StepJEPA \textbf{removes that ceiling by allocating one latent block per logical step} and learning a recurrent world-model-esque predictor. As width and depth increase, the encoder packs richer deductions into each latent block while the predictor accumulates less rollout drift, making multi-task training on billion-parameter backbones the natural next step.

\section{Discussion and Conclusion}
\label{sec:discussion}
We proposed and evaluated a step-level formulation for latent reasoning: encoding each logical step into a compact latent block and directly supervising step-to-step transitions via JEPA, SIGReg, joint step decoding, and adaptive halting inside one causal transformer. Rather than training unconstrained hidden states indirectly or distilling a fixed budget at a single anchor, StepJEPA learns an explicit recurrent transition predictor in representation space while keeping imagined steps readable at test time.

More broadly, showing that 19M autoregressive models acquire transferable multi-step dynamics where explicit CoT and distillation remain at chance suggests that step-level joint-embedding prediction may unlock structured reasoning in far smaller models. Promising next steps include multi-task training on larger backbones, test-time compute scaling over passes $M$ and Jacobi iterations $I$, and bidirectional or diffusion extensions.

\section*{Limitations}
Four practical limitations merit note (\cref{app:neg}): (i) StepJEPA currently relies on step-segmented training traces, though automatic segmentation can relax this requirement; (ii) multi-role encoding and multi-step self-rollouts increase training cost substantially relative to standard fine-tuning (\cref{tab:app_wallclock}), even though inference is fast; (iii) deterministic MSE prediction can average across valid alternative next steps on open-ended word problems \citep{dinh2026jepaparadox}, where fixed-budget PCCoT still leads in-distribution GSM8k; and (iv) very deep out-of-distribution rollouts can still drift or saturate (as seen on 19M cellular automata, Long Mazes, and 9--10 hop Deduction), where learned halting heads remain less reliable than explicit state-based halting.

\bibliography{refs}
\bibliographystyle{icml2026}

\newpage
\appendix
\onecolumn

\section{Extended Related Work, Paradigm Comparison, SIGReg Form, and Algorithm}
\label{app:method}
To situate StepJEPA transparently across the rapidly expanding 2024--2026 literature (summarized in recent surveys; \citealp{chen2025latentsurvey,zhu2025latentsurvey}), \cref{tab:related} contrasts all major latent-reasoning, recurrent-depth, concept-space, and language-JEPA lineages. Due to computational constraints, we cannot re-implement and retrain every concurrent paradigm across every benchmark for this initial release; instead, we benchmark directly against canonical end-to-end baselines (Explicit CoT, Coconut, CODI, PCCoT, Direct Answer) and use our matched No-JEPA control ($\lambda_J=0$, retaining joint step-text decoding, complemented by an encoder-free self-rollout variant to eliminate gold-encoder input mismatch) to isolate the effect of $\mathcal{L}_{\text{JEPA}}$.

\begin{table}[ht]
\centering
\caption{Expanded comparison across latent-reasoning, recurrent-depth, concept-space, and language-JEPA paradigms. StepJEPA is the only framework combining logical-step latent blocks, single-stage deterministic stop-gradient JEPA + SIGReg transition prediction, a single unified causal backbone, and adaptive halting.}
\label{tab:related}
\resizebox{\textwidth}{!}{%
\begin{tabular}{lllll}
\toprule
Method & Latent Unit & Supervision / Transition Signal & Unified Backbone? & Adaptive Depth? \\
\midrule
Coconut \citep{hao2024coconut} & Last hidden state & Indirect LM loss on following text & Yes & No \\
CODI \citep{shen2025codi} & Fixed cont.\ tokens (6 default) & Single-position teacher KD & Yes & No \\
PCCoT \citep{wu2025pccot} & Fixed Jacobi tokens (24 default) & Single-position teacher KD (Jacobi) & Yes & No \\
SIM-CoT \citep{wei2025simcot} & Per-step implicit token & Auxiliary step-text decoder CE (no JEPA) & Separate aux.\ decoder & No (fixed budget) \\
PLaT \citep{wang2026plat} & Continuous planner state & Decoupled decoder rec.\ CE (no JEPA) & Separate planner/dec. & Yes (answer-aware) \\
LaDiR \citep{kang2025ladir} & VAE step thought block & 2-stage $\beta$-VAE + blockwise latent diffusion & Separate VAE + diffuser & Yes (binary head) \\
CoLaR \citep{tan2025colar} & Merged $c$-token chunk & Next-compressed-embedding head + RL & Yes & Prompted factor $c$ \\
Huginn / Ouro \citep{geiping2025huginn,zhu2025ouro} & Per-token vertical state & NTP (Huginn) / ent.-reg.\ NTP + exit (Ouro) & Custom looped pretrain & Yes (per-token exit) \\
LCM / CALM \citep{barrault2024lcm,shao2025calm} & Sentence / $K$-tok chunk & MSE/diff.\ (LCM); Energy Score (CALM) & External SONAR / AE & No (per sent./chunk) \\
LLM-JEPA / STP \citep{llmjepa2025,huang2026stp} & View / token-span state & Auxiliary JEPA / cosine tube; text inference & Yes & No (text decoding) \\
SSP \citep{yuan2026ssp} & Step-delimiter state & Step-boundary STP + post-hoc probe; text inf. & Yes (no predictor) & No (text decoding) \\
JEPA-Reasoner \citep{jepareasoner2025} & Token hypersphere latent & Scaled cosine to EMA token emb. & Separate Talker (frozen Reas.) & No \\
\textbf{StepJEPA (ours)} & \textbf{Logical-step block ($K$ slots)} & \textbf{Closed-loop step JEPA + SIGReg + joint dec.} & \textbf{Yes ($<1\%$ overhead)} & \textbf{Yes (head / state)} \\
\bottomrule
\end{tabular}}
\end{table}

\textbf{Extended lineage discussion.}

\emph{(1) Implicit, discrete-token, and fixed-budget continuous CoT:} Implicit CoT internalizes reasoning without horizontal latent tokens via vertical layer-wise distillation \citep{deng2023icotkd} or progressive CoT token removal \citep{deng2024icot}, while discrete compute tokens append learnable pause tokens \citep{goyal2023pause} or meaningless filler dots requiring dense supervision \citep{pfau2024filler}. Fixed-budget continuous-CoT methods instead feed continuous hidden states horizontally, trained via downstream LM loss (Coconut; \citealp{hao2024coconut}) or single-position all-layer teacher self-distillation (CODI \citealp{shen2025codi}; parallel Jacobi PCCoT \citealp{wu2025pccot}). Mechanistic and causal audits show that such trajectories can behave as deletable placeholders or prompt shortcuts \citep{zhang2025dolatent,qiu2026actuallyreason} or lose faithful step-by-step rollouts as reasoning depth grows \citep{liang2026mechanistic}, directly motivating StepJEPA's explicit step-transition supervision and causal perturbation checks.

\emph{(2) Step-supervised, compressed, hybrid, and multi-stage latent CoT:} Coconut+SIM-CoT \citep{wei2025simcot} (44.8\% published on GPT-2 GSM8k-Aug; +4.3 pts OOD math average over Coconut) and PLaT \citep{wang2026plat} (targeting Pass@$k$ diversity on GPT-2 and Llama-3.2-1B/3B) align latent states to explicit steps purely via auxiliary or decoupled text-reconstruction decoders without a step encoder or representation-space JEPA loss; our No-JEPA control ($\lambda_J=0$) isolates why step-text decoding alone still fails from-scratch grokking and deep OOD extrapolation. Multi-stage codec frameworks include LaDiR \citep{kang2025ladir} (two-stage $\beta$-VAE with blockwise latent diffusion) and FLaRe \citep{ouali2026flare} (whole-CoT VAE with flow matching and verified self-training). Compression and alignment variants include CoLaR \citep{tan2025colar} (+14.1 pts avg over Coconut via Stage-1 SFT dynamic $c$-token embedding merging, plus up to +5.4\% on MATH via RL), CCoT \citep{cheng2024ccot} (approximating a learned-scorer subset of gold CoT hidden states), KaVa \citep{kuzina2025kava} (compressed teacher KV-cache distillation), PMPS \citep{li2026pmps} (learnable prototype alignment), and Supervised Thinking States \citep{amos2026thinkingstates} (interleaved thinking tokens during input ingestion). Complementary vocabulary and hybrid lines include discrete VQ-VAE latent tokens (Token Assorted; \citealp{su2025tokenassorted}), supervised vocabulary superpositions (Latent-SFT; \citealp{deng2025latentsft}), training-free probability-weighted soft embeddings (Soft Thinking; \citealp{zhang2025softthinking}), and RL-gated hidden/token blending (HRPO; \citealp{yue2025hrpo}).

\emph{(3) Recurrent-depth, looped transformers, recursive puzzle solvers, and adaptive halting:} Vertically looped transformers iterate shared layers per token (Huginn \citealp{geiping2025huginn}, \citealp{saunshi2025looped}, Ouro \citealp{zhu2025ouro}, and algorithmic length-generalizing looped transformers \citealp{fan2024loopedlength}) or iteratively refine fixed latent workspaces (LOTUS \citealp{fan2026lotus}, LLoCoT \citealp{garcia2026llocot}). Separate specialized architectures excel on grid puzzles: HRM \citep{wang2025hrm} and TRM \citep{jolicoeur2025trm} (7M parameters, 85.3\% on fixed-size 30$\times$30 Maze-Hard from scratch) solve hard fixed-grid puzzles from few examples, while Recall \citep{bansal2022recall} achieves spatial maze-size extrapolation via recurrent residual networks. StepJEPA is orthogonal: it operates horizontally over logical-step latents inside a standard causal LM (compatible with off-the-shelf GPT-2 via LoRA) and combines adaptive horizontal halting \citep{graves2016act,dehghani2019ut,banino2021pondernet,ning2025stop} with inner $M$-pass draft refinement.

\emph{(4) Concept-space LMs, JEPA world models, and length generalization:} For general text generation, Large Concept Models \citep{barrault2024lcm} predict frozen SONAR sentence embeddings (MSE or diffusion), CALM \citep{shao2025calm} predicts autoencoder-compressed $K$-token chunks via a single-step Energy Score head, and DLCM \citep{qu2025dlcm} dynamically pools variable-length concepts inside a hierarchical NTP transformer. In JEPA world modeling, LeWorldModel \citep{maes2026lewm} and \citet{klindt2026lejepawm} establish SIGReg-based latent prediction and linear identifiability, while \citet{dinh2026jepaparadox} analyze centroid collapse in unconditioned masked-text MSE JEPA. For language, LLM-JEPA \citep{llmjepa2025}, STP \citep{huang2026stp} (cosine-displacement regularization across random token triples), DLLM-JEPA \citep{dllmjepa2026}, and NextLat \citep{teoh2025nextlat} (auxiliary token-level dynamics discarded at test time) retain explicit token-level inference, whereas VL-JEPA \citep{chen2025vljepa}, JEPA-Reasoner \citep{jepareasoner2025} (separate Talker over a frozen token-level Reasoner), and JELAR \citep{wu2026jelar} (fixed-$K$ audio-LALM latent reasoning via frozen WavJEPA and BERT) rely on separate or frozen modules without logical-step halting. Unlike STP's random-token straightening or SSP's step-boundary smoothing and offline teacher-forced probing \citep{yuan2026ssp} (which reports no GSM8k gain over vanilla fine-tuning and chance error detection), StepJEPA learns a problem-conditioned neural transition model $P(x, z_{0:t}) \to (\hat{z}_{t+1}, p^{\text{done}}_{t+1})$ over canonicalized, SIGReg-regularized step blocks and executes closed-loop, adaptive-depth reasoning strictly in latent space. Finally, our OOD length-extrapolation design connects to transformer expressivity and positional-generalization studies \citep{merrill2024cotpower,zhou2024raspl,kazemnejad2023nope,kiyono2021shape,ruoss2023randomized,mcleish2024abacus,cho2024poscoupling}, showing that task-agnostic positional jitter plus step-level JEPA transitions enable strong depth extrapolation without task-specific digit-alignment embeddings.

\textbf{Step encoder attention pooling.} Given transformer hidden states $h^{\text{enc}}_{1:L_t}$ for logical step $s_t$ and $K$ learned queries $q_1, \dots, q_K \in \mathbb{R}^d$, each canonicalized slot vector is
\begin{equation}
  z_{t,k} = \LN\!\left(\sum_{\ell=1}^{L_t} \alpha_{k,\ell} h^{\text{enc}}_\ell\right), \qquad
  \alpha_{k,\ell} = \operatorname{softmax}_\ell\!\left(\frac{q_k^\top h^{\text{enc}}_\ell}{\sqrt{d}}\right), \qquad k = 1, \dots, K.
\end{equation}

\textbf{Exact SIGReg quadrature.} Following LeJEPA \citep{balestriero2025lejepa}, given a batch of $N$ latent slot vectors $\{v_n\}_{n=1}^N \subset \mathbb{R}^d$, we sample $D_u = 256$ random unit directions $u_j \in \mathbb{S}^{d-1}$, project $a_{n,j} = v_n^\top u_j$, and minimize the Epps--Pulley distance between the empirical characteristic function and that of $\mathcal{N}(0, 1)$ over a 17-point quadrature grid $t \in [-4, 4]$:
\begin{equation}
  \mathcal{L}_{\text{SIGReg}}(v) = \frac{1}{D_u}\sum_{j=1}^{D_u}\int_{-4}^{4} \Bigl|\frac{1}{N}\sum_{n=1}^{N} e^{i t a_{n,j}} - e^{-t^2/2}\Bigr|^2 e^{-t^2/2}\,dt.
\end{equation}

\textbf{Halting binary cross-entropy.} With binary stopping targets $y^{\text{done}}_t = \mathbb{I}[t = T]$, the halting head is trained via
\begin{equation}
  \mathcal{L}_{\text{done}} = -\frac{1}{T}\sum_{t=1}^{T}\Bigl(y^{\text{done}}_t \log p^{\text{done}}_t + (1 - y^{\text{done}}_t)\log(1 - p^{\text{done}}_t)\Bigr).
\end{equation}

\textbf{Auxiliary explicit-CoT cross-entropy.} Writing $u = (s_1, \dots, s_T, y, \langle\text{eos}\rangle)$ for the concatenated written reasoning steps and final answer (capped at width 128 by default, normalized per valid token across the batch), the optional training-only CoT loss is
\begin{equation}
  \mathcal{L}_{\text{cot}} = -\frac{1}{L} \sum_{j=1}^{L} \log p_\theta\!\left(u_j \mid x, u_{<j}\right).
\end{equation}

\textbf{Parallel deterministic Jacobi update and training pass summation.} For the parallel Jacobi variant, we initialize a fixed horizon of $H$ latent blocks with a learned null embedding $Z^{(0)}_{1:H} = \LN(d_{\text{jac}})$ and perform $I$ parallel block-causal fixed-point updates using the same draft-slot layout: at iteration $i = 1, \dots, I$, block $t$'s committed input holds $Z^{(i-1)}_{t-1}$ (with $Z^{(i-1)}_0 = z_0$) while its draft input holds $Z^{(i-1)}_t$, yielding $Z^{(i)}_{1:H} = P(x, z_0, Z^{(i-1)}_{1:H})$. During training, $\mathcal{L}_{\text{JEPA}}$ supervises the final iterate $Z^{(I)}$ (and optionally intermediate iterates with weight $w_{\text{ref}}$) with end-to-end gradients through all $I$ iterations. Across training passes, the total objective sums the base teacher-forced terms, the final rollout-pass terms ($s = S_{\text{ro}}$ for $\mathcal{L}_{\text{JEPA}}, \mathcal{L}_{\text{pdec}}, \mathcal{L}_{\text{done}}, \mathcal{L}_{\text{ans}}$, plus $\mathcal{L}_{\text{SIGReg}}(\hat{z})$ when enabled), and any scheduled-sampling pass ($p_{\text{ro}} > 0$), while $\mathcal{L}_{\text{rec}}$, $\mathcal{L}_{\text{SIGReg}}(z)$, and gold-readout mixing $\alpha_g$ are evaluated once on the base pass and $\lambda_J$ ramps linearly over the first 3,000 steps.

\begin{algorithm}[ht]
\caption{StepJEPA Training Step and Adaptive Sequential Inference}
\label{alg:stepjepa}
\begin{algorithmic}[1]
\small
\STATE \textbf{Training} (problem $x$, steps $s_{0:T}$, answer $y$; passes $M$, rollouts $S_{\text{ro}}$):
\STATE Encode clean targets $z_t \leftarrow E(s_t)$ for $t = 0, \dots, T$ (no dropout).
\STATE \textbf{for} refinement pass $m = 1 \dots M$ \textbf{do}
\STATE \quad Draft $\hat{z}^{(m)}_{1:T}, p^{\text{done}}_{1:T} \leftarrow P(x, z_{0:T-1}; \text{draft}^{(m-1)})$ (parallel across $t$).
\STATE \quad Update $\text{draft}^{(m)} \leftarrow W_{\text{in}} \hat{z}^{(m)}_{1:T} + e_{\text{draft}}$.
\STATE \textbf{end for}; set $\hat{z}^{\text{base}} \leftarrow \hat{z}^{(M)}$, $\hat{z}^{\text{ro}} \leftarrow \hat{z}^{\text{base}}$.
\STATE \textbf{for} rollout $s = 2 \dots S_{\text{ro}}$ \textbf{do}
\STATE \quad $\hat{z}^{\text{ro}}, p^{\text{done}} \leftarrow P(x, [z_0, \hat{z}^{\text{ro}}_{1:T-1}])$ ($M$ refinement passes, end-to-end grad).
\STATE \textbf{end for}
\STATE Sum $\mathcal{L}_{\text{StepJEPA}}$ (\cref{eq:loss}, $\sg(z_t)$ on JEPA targets) across $\hat{z}^{\text{base}}$, $\hat{z}^{\text{ro}}$, and optional detached scheduled-sampling pass; update $\theta$.
\vspace{0.35em}
\STATE \textbf{Adaptive Sequential Inference} (problem $x$, initial $s_0$, threshold $\tau$):
\STATE $z_0 \leftarrow E(s_0)$; $\mathcal{H} \leftarrow [z_0]$.
\STATE \textbf{for} $t = 1 \dots T_{\max}$ \textbf{do}
\STATE \quad Run $M$ draft-and-refine passes of $P(x, \mathcal{H})$ to get $\hat{z}_t, p^{\text{done}}_t$.
\STATE \quad Append $\hat{z}_t$ to $\mathcal{H}$; \textbf{if} $p^{\text{done}}_t > \tau$ \textbf{then break}
\STATE \textbf{end for}
\STATE Emit final answer $\hat{y} \leftarrow R(x, \hat{z}_{1:t})$ autoregressively; \textbf{return} $\hat{y}$.
\end{algorithmic}
\end{algorithm}

\section{Benchmark Suite and ProntoQA Non-Overfitting Diagnostics}
\label{app:tasks}
\label{app:prontoqa}
We evaluate StepJEPA across standard language-model reasoning benchmarks (\textbf{GSM8k-Aug}, \textbf{MultiArith}, \textbf{SVAMP}, \textbf{GSM-Hard}, \textbf{StrategyQA-CoT}, \textbf{ProsQA}, \textbf{ProntoQA}, and \textbf{StepGame}) together with four newly constructed procedural out-of-distribution (OOD) length-extrapolation benchmarks designed to isolate reasoning depth from surface shortcuts and unseen-token confounds; all four procedural generators and splits will be released alongside our code. Every dataset is partitioned into strictly disjoint training, validation, and test splits verified to have zero exact or normalized problem overlap.

\begin{itemize}
  \item \textbf{GSM8k-Aug} \citep{cobbe2021gsm8k,deng2024icot}: Grade-school arithmetic word problems with equation-level step traces (385,620 train / 500 validation / 1,319 unseen test problems), scored by exact numeric match following CODI and Coconut.
  \item \textbf{ProsQA} \citep{hao2024coconut}: Multi-hop directed-graph reachability expressed over fictional entity hierarchies (17,886 train / 300 validation / 500 unseen test problems), testing structured search over branching ontologies.
  \item \textbf{StrategyQA-CoT} \citep{geva2021strategyqa}: Implicit multi-hop commonsense yes/no reasoning benchmark formatted with CODI's chain-of-thought splits (1,609 train / 200 validation / 229 unseen test problems).
  \item \textbf{ProntoQA} \citep{saparov2023prontoqa}: Deductive logical syllogisms requiring multi-step rule chaining to determine truth values (9,000 train / 200 validation / 800 unseen test problems).
  \item \textbf{Mazes and Long Mazes (new procedural benchmark)}: Newly constructed text-encoded shortest-path planning benchmark inspired by algorithmic maze suites \citep{bansal2022recall,jolicoeur2025trm}, featuring randomized dead ends, loop passages, and step-by-step shortest-path traces within a fixed coordinate vocabulary ($4\times 4$--$6\times 6$ grids for 164M AR, trained on 480,000 online-streamed mazes / 256 validation / 500 unseen test per split; $4\times 4$--$10\times 10$ grids for 19M AR Long Mazes, 100,000 train / 512 validation / 500 unseen test per split). Training examples contain shortest paths of 4--12 cells; unseen OOD test splits evaluate paths of 13--15 and 16--20 cells (up to $1.67\times$ training length, 164M AR) and 21--28 and 29--40 cells (up to $3.33\times$ training length, 19M AR).
  \item \textbf{Multi-Hop Deduction (new procedural benchmark)}: Newly constructed shortcut-free logical-chaining benchmark inspired by ProntoQA \citep{saparov2023prontoqa} and ProofWriter \citep{tafjord2021proofwriter}, explicitly balancing positive and negated rules with shared-premise distractors so surface polarity and frequency cues remain at chance (100,000 train / 512 validation, 300 evaluated per checkpoint / 500 unseen test problems per split); trained and ID-tested on 2--6 reasoning hops and tested on unseen OOD1 (7--8 hops) and OOD2 (9--10 hops) chains.
  \item \textbf{One-Dimensional Cellular Automata (CA1D; new procedural benchmark)}: Newly constructed text-simulation benchmark of elementary 1D Wolfram cellular automata (rules 0--255; \citealp{wolfram1983ca}) over fixed-width binary rows (16--24 bits) with unary step-count encoding (100,000 train / 512 validation, 300 evaluated per checkpoint / 500 unseen test problems per split); trained and ID-tested on 1--4 rollout steps and tested on unseen OOD1 (6 steps) and OOD2 (8 steps) horizons without introducing unseen step tokens.
  \item \textbf{iGSM-Small Arithmetic DAGs (new procedural variant)}: Newly constructed compact procedural variant inspired by iGSM \citep{ye2024igsm}, formulating shuffled arithmetic variable-dependency DAGs with decoy definitions and exact non-negative small-integer arithmetic (rather than modular residues) so every instance is uniquely verifiable from its prompt text (100,000 train / 512 validation, 300 evaluated per checkpoint / 500 unseen test problems per split); trained and ID-tested on 3--6 operations and tested on strictly deeper unseen OOD1 (8 operations) and OOD2 (10 operations) expression graphs.
  \item \textbf{StepGame} \citep{shi2022stepgame}: Multi-hop spatial reasoning tracking relative cardinal directions across entities (100,000 train / 512 validation, 300 evaluated per checkpoint / 500 unseen test problems per split), trained on 2--6 spatial hops and tested on 4-hop (ID), 8-hop (OOD1), and 10-hop (OOD2) splits (audited for prompt shortcuts as noted in \cref{app:neg}).
\end{itemize}
Across every task, the initial step $s_0$ contains no depth or gold-answer metadata: it holds the explicit starting state on state-tracking tasks (starting cell on Mazes, initial bit row on CA1D), the prompt-stated target variable query on iGSM-Small (\texttt{question : what is <var> ?}), the prompt-stated starting entity on StepGame, and a constant neutral marker (\texttt{start .}) on GSM8k, Multi-Hop Deduction, ProntoQA, ProsQA, and StrategyQA-CoT.

\textbf{Ruling out baseline overfitting on from-scratch 19M ProntoQA.} \Cref{tab:prontoqa_val} summarizes the full 40-epoch validation trajectory and final training loss for our matched 19M from-scratch autoregressive models on ProntoQA. Because checkpoints are selected strictly by per-epoch validation accuracy across all 40 epochs, the failure of Explicit CoT, CODI, PCCoT, No-JEPA, and Encoder-Free No-JEPA at $\approx$50\% is not caused by over-training past a good checkpoint: none ever rose above chance validation accuracy at any epoch, even as training loss dropped smoothly. StepJEPA alone underwent a clean grokking transition from 52.0\% (step 1,775) to 82.0\% (step 1,988) reaching 100.0\% validation accuracy (first reached at step 2,201; selected tied-peak checkpoint at step 2,272), achieving 99.25\% on the 800-problem unseen test set.

\begin{table}[ht]
\centering
\caption{40-epoch validation trajectory and training-loss summary for matched 19M AR models trained from random initialization on ProntoQA ($n=800$ unseen test). Every method was evaluated every epoch (40 checkpoints) and selected strictly by peak validation accuracy; best accuracy per column is \textbf{bold} and second best is \textit{italic}.}
\label{tab:prontoqa_val}
\small
\begin{tabular}{lcccc}
\toprule
Method (19M AR From Scratch) & Val Min--Max (\%) $\uparrow$ & Best Step & Final Train Loss $\downarrow$ & Unseen Test (\%) $\uparrow$ \\
\midrule
Explicit CoT & 42.0--56.5 & 1,065 & 0.0014 & 50.00 \\
CODI & 41.5--55.5 & 1,917 & 0.5885 & 50.38 \\
PCCoT & \textit{44.5--58.0} & 1,846 & 0.5247 & 50.13 \\
No-JEPA Control ($\lambda_J = 0$) & 44.5--57.5 & 710 & 0.1446 & \textit{50.75} \\
Encoder-Free Self-Rollout No-JEPA & \textit{44.5--58.0} & 1,491 & 0.4004 & 50.00 \\
\textbf{StepJEPA (ours)} & \textbf{44.5--100.0} & 2,272 & 0.1499 & \textbf{99.25} \\
\bottomrule
\end{tabular}
\end{table}

\section{Decoded Thoughts, Causal Ablations, and Component Matrix}
\label{app:halting}
\label{app:qualitative}
\label{app:composability}
\label{app:v3}
\textbf{Decoded imagined thoughts.} \Cref{tab:qualitative} displays verbatim step-decoder reconstructions $D(\hat{z}_t)$ of internal imagined latents from our 124M GPT-2 StepJEPA checkpoint across in-distribution GSM8k and zero-shot MultiArith and SVAMP.

\begin{table}[ht]
\centering
\caption{\textbf{Inspectable latent thoughts decoded via the shared step decoder $D(\hat{z}_t)$} on 124M GPT-2 StepJEPA (trained solely on GSM8k, evaluated zero-shot on MultiArith and SVAMP). At inference time, StepJEPA emits only the final answer; passing internal imagined latents through $D$ reveals readable arithmetic steps and adaptive early halting, while Row 4 illustrates that on off-manifold error trajectories, $D(\hat{z}_4)$ (\texttt{24}) and the readout $R$ (\texttt{23}) can diverge.}
\label{tab:qualitative}
\scriptsize
\setlength{\tabcolsep}{4pt}
\begin{tabularx}{\textwidth}{@{}l X X c c@{}}
\toprule
Benchmark & Problem Excerpt & Decoded Imagined Latent Steps $D(\hat{z}_1), \dots, D(\hat{z}_{\hat{T}})$ & Steps / NFE & Pred / Gold \\
\midrule
GSM8k (ID) & Toulouse has $2\times$ as many sheep as Charleston, who has $4\times$ Seattle's 20 sheep. Total sheep? &
  $\hat{z}_1$: \texttt{<<4*20=80>>} $\;\to\;$ $\hat{z}_2$: \texttt{<<2*80=160>>} $\;\to\;$ $\hat{z}_3$: \texttt{<<80+160+20=260>>} &
  3 / 9 & \textcolor{win}{\textbf{260}} / 260 \\
MultiArith (ZS) & Paige had 11 songs, deleted 9 old songs, and added 8 new songs. How many songs now? &
  $\hat{z}_1$: \texttt{<<11-9=2>>} $\;\to\;$ $\hat{z}_2$: \texttt{<<2+8=10>>} \quad \textit{(halts early)} &
  2 / 7 & \textcolor{win}{\textbf{10}} / 10 \\
SVAMP (ZS) & Olivia had \$78 in her wallet and spent \$15 at a supermarket. How much money is left? &
  $\hat{z}_1$: \texttt{<<78-15=63>>} \quad \textit{(halts after 1 logical step)} &
  1 / 5 & \textcolor{win}{\textbf{63}} / 63 \\
GSM8k (Error) & One glass costs \$5, every second glass costs 60\% of the price. Total cost for 16 glasses? &
  $\hat{z}_1$: \texttt{<<5*0.6=3>>} $\;\to\;$ $\hat{z}_2$: \texttt{<<16*5=80>>} $\;\to\;$ $\hat{z}_3$: \texttt{<<16*3=48>>} $\;\to\;$ $\hat{z}_4$: \texttt{<<16+8=24>>} &
  4 / 11 & \textcolor{base}{23} / 64 \\
\bottomrule
\end{tabularx}
\end{table}

\textbf{Pure-autoregressive step-decodability ablation across scales.} As noted in \cref{sec:arch}, a strict pure-JEPA view might treat the step decoder $D$ purely as an external interpretability probe whose gradients are blocked from touching the encoder and predictor. \Cref{tab:decodability} tests this hypothesis on our matched pretrained autoregressive SMDM-AR backbones: full joint training leads at both 19M (\textbf{26.69\%} vs.\ 26.23\% detached and 26.38\% no-decoder, with 98.1\%--98.5\% exact-match step reconstruction) and 48M (\textbf{32.22\%} vs.\ 32.15\% detached, 31.99\% no-decoder, and 31.46\% reduced), while detached probing leads at 34M (\textbf{29.87\%} vs.\ 28.96\% no-decoder, 28.73\% reduced, and 28.66\% full-joint), confirming that $\mathcal{L}_{\text{JEPA}}$ itself carries the primary reasoning signal while joint decoding provides readable thoughts at negligible or positive accuracy cost.

\begin{table}[ht]
\centering
\caption{Pure-autoregressive step-decodability ablation on GSM8k unseen test ($n=1,319$, matched SlimPajama-pretrained SMDM-AR backbones, accuracy in \%; best per row is \textbf{bold}, second best is \textit{italic}; 48M No Decoder completed epoch 10 via a 1-epoch continuation from its epoch-9 checkpoint at peak learning rate $10^{-4}$).}
\label{tab:decodability}
\small
\begin{tabular}{lcccc}
\toprule
Model Scale (AR; Acc.\ \% $\uparrow$) & Full Joint ($w=1$) & Reduced ($w=0.3$) & Detached Probe & No Decoder \\
\midrule
19M AR & \textbf{26.69} & 22.82 & 26.23 & \textit{26.38} \\
34M AR & 28.66 & 28.73 & \textbf{29.87} & \textit{28.96} \\
48M AR & \textbf{32.22} & 31.46 & \textit{32.15} & 31.99 \\
\bottomrule
\end{tabular}
\end{table}

\textbf{Controlled No-JEPA ($\lambda_J=0$) causal ablation table.} \Cref{tab:jepa_nec} collects the full matched No-JEPA ablation across architectures, scales, and pretraining regimes discussed in \cref{sec:jepa_mech}.

\begin{table}[ht]
\centering
\caption{Controlled ablation removing \emph{only} $\mathcal{L}_{\text{JEPA}}$ ($\lambda_J = 0$) with identical architecture, data, latent loop, step decoder, and readout across unified strictly causal AR backbones (19M--164M; unseen test accuracy in \%; higher per split is \textbf{bold}, ties go to StepJEPA; \dag strictly encoder-free self-rollout No-JEPA control with zero gold-latent teacher forcing).}
\label{tab:jepa_nec}
\small
\resizebox{\textwidth}{!}{%
\begin{tabular}{llcc}
\toprule
Model Setting (Strictly Causal AR) & Unseen Test Split (Acc.\ \% $\uparrow$) & \textcolor{ours}{\textbf{StepJEPA}} & No-JEPA ($\lambda_J=0$) \\
\midrule
19M AR From Scratch (ProntoQA) & Unseen test ($n=800$) & \textbf{99.25} & 50.75 (50.00\dag) \\
19M AR From Scratch (iGSM-Small) & ID / OOD1 / OOD2 & \textbf{96.0 / 26.0 / 16.4} & 78.0 / 17.4 / 12.2 (39.2 / 10.8 / 9.8\dag) \\
19M AR From Scratch (CA1D) & ID unseen test & \textbf{94.6} & 83.8 \\
19M AR From Scratch (Long Mazes) & ID (7--12) / OOD (21--40) & \textbf{53.8} / 0.0 & 0.0 / 0.0 \\
19M Pretrained AR (GSM8k) & Unseen test ($n=1,319$) & \textbf{26.69} & 18.88 \\
124M GPT-2 LoRA (ProsQA) & Unseen test ($n=500$) & 94.4 & \textbf{95.0} \\
124M GPT-2 LoRA (StrategyQA-CoT) & Unseen test ($n=229$) & \textbf{60.26} & 53.28 \\
124M GPT-2 LoRA (iGSM-Small) & ID / OOD1 / OOD2 & \textbf{89.6 / 60.2 / 35.2} & 73.8 / 27.0 / 18.6 \\
124M GPT-2 LoRA (StepGame) & 4-hop ID / 8 / 10-hop OOD & \textbf{99.4} / 95.6 / \textbf{90.0} & 99.4 / \textbf{96.2} / 55.6 \\
124M GPT-2 LoRA (Multi-Hop Deduction) & ID / OOD1 / OOD2 & \textbf{99.8 / 73.4} / 47.2 & 87.0 / 69.0 / \textbf{64.0} \\
164M AR From Scratch (Mazes, 3 seeds) & ID (4--12) / OOD (13--15, 16--20) & \textbf{98.6} / \textbf{90.1 / 74.2} & 3.2 / 0.0 / 0.0 \\
164M AR Pretrained (Mazes, 3 seeds) & ID (4--12) / OOD (13--15, 16--20) & \textbf{83.9} / \textbf{29.9 / 26.8} & 6.3 / 0.0 / 0.0 \\
\bottomrule
\end{tabular}}
\end{table}

\textbf{Decoding schedules and CoT composability.} Two additional controlled comparisons clarify how StepJEPA interacts with decoding schedules and auxiliary supervision:
\begin{itemize}
  \item \textbf{Adaptive sequential vs.\ parallel Jacobi decoding.} On 124M GPT-2 GSM8k, adaptive sequential refinement maximizes accuracy (\textbf{44.96\%} in 9.90 per-example NFE, 34.3s batched total), whereas parallel deterministic Jacobi trades 3.3 points for ultra-low constant-horizon latency (\textbf{41.62\%} in 7.22 per-example NFE, 20.6s batched total; zero-shot transfer reaches 93.9\% MultiArith in 7.01 NFE, 45.2\% SVAMP in 7.17 NFE, and 9.48\% GSM-Hard in 8.50 NFE). Both use the exact same StepJEPA objective.
  \item \textbf{Orthogonality with auxiliary CoT.} On our pure-AR 19M ablation (\cref{tab:app_v3}), removing auxiliary text CoT ($w_{\text{cot}}=0$) costs only $-0.76$ percentage points (26.69\% $\to$ 25.93\%), whereas removing $\mathcal{L}_{\text{JEPA}}$ ($\lambda_J=0$) collapses accuracy by $-7.81$ points (26.69\% $\to$ 18.88\%), confirming that JEPA rather than auxiliary CoT drives performance (controlled GSM8k loss removals are performed on the matched 19M AR backbone in \cref{tab:app_v3}).
\end{itemize}

\textbf{Extended pure-AR component ablation matrix.} \Cref{tab:app_v3} reports the full 13-run single-addition, leave-one-out, and loss-removal ablation matrix on our matched pretrained 19M autoregressive backbone (SMDM-AR) evaluated on the full GSM8k unseen test set ($n=1,319$).

\begin{table}[ht]
\centering
\caption{Pure-autoregressive component ablation matrix on GSM8k unseen test ($n=1,319$, pretrained 19M SMDM-AR backbone; best is \textbf{bold}, second best is \textit{italic}).}
\label{tab:app_v3}
\small
\resizebox{\textwidth}{!}{%
\begin{tabular}{llccc}
\toprule
Group & Variant (19M Pretrained AR) & Test Acc.\ (\%) $\uparrow$ & $\Delta$ vs.\ Base $\uparrow$ & $\Delta$ vs.\ Comb.\ $\uparrow$ \\
\midrule
Baseline & Standard AR StepJEPA (base) & 24.03 & 0.00 & $-2.66$ \\
\midrule
Single addition & + Isotropy reg.\ on predicted latents ($\mathcal{L}_{\text{SIGReg}}(\hat{z})$) & 24.87 & +0.84 & $-1.82$ \\
Single addition & + Deeper multi-step self-rollout ($S_{\text{ro}}=3, p=0.35$) & 24.56 & +0.53 & $-2.13$ \\
Single addition & + No step emb.\ + positional jitter ($J=16$) & 23.73 & $-0.30$ & $-2.96$ \\
Single addition & + Mixed gold/predicted readout ($\alpha_g=0.5$) & 23.12 & $-0.91$ & $-3.57$ \\
\midrule
Combined & Combined refined objective (all 4) & \textit{26.69} & \textit{+2.66} & \textit{0.00} \\
Leave-one-out & \textbf{Combined $-$ mixed gold/predicted readout ($\alpha_g=0$)} & \textbf{26.91} & \textbf{+2.88} & \textbf{+0.22} \\
Leave-one-out & Combined $-$ no step emb.\ / jitter & 24.64 & +0.61 & $-2.05$ \\
Leave-one-out & Combined $-$ predicted-latent SIGReg & 24.34 & +0.31 & $-2.35$ \\
Leave-one-out & Combined $-$ deep self-rollout & 21.76 & $-2.27$ & $-4.93$ \\
\midrule
Loss removal & Combined without auxiliary CoT ($w_{\text{cot}}=0$) & 25.93 & +1.90 & $-0.76$ \\
Loss removal & Combined without SIGReg ($\mathcal{L}_{\text{SIGReg}}=0$) & 23.43 & $-0.60$ & $-3.26$ \\
Loss removal & Combined without JEPA ($\lambda_J=0$, No-JEPA) & 18.88 & $-5.15$ & $-7.81$ \\
\bottomrule
\end{tabular}}
\end{table}

\section{In-Domain Capacity Scaling and Zero-Shot Transfer}
\label{app:id_scaling}
\Cref{tab:scaling} reports our autoregressive scaling ladder across matched SlimPajama-pretrained SMDM-AR backbones (19M, 34M, 48M) and 124M GPT-2 on in-domain GSM8k ($n=1,319$), zero-shot OOD arithmetic transfer (MultiArith / SVAMP), and iGSM-Small math DAG depth extrapolation. Explicit CoT at 19M--48M is evaluated with an untruncated 256-token budget (necessary because TinyLlama digit-level tokenization averages 40--139 CoT tokens). Notice that at 19M--48M, untruncated Explicit CoT benefits from 40--139 tokens of external digit-by-digit scratchpad compute per problem, whereas StepJEPA compresses each arithmetic equation into $K=2$ latent slots; consequently, StepJEPA already surpasses both CODI (19.26\% / 21.00\% / 24.34\%) and PCCoT (19.33\% / 22.29\% / 26.84\%) across every compact scale (27.22\% / 29.87\% / 32.15\% GSM8k, and 63.9\% $\to$ 75.0\% $\to$ 83.9\% MultiArith) and overtakes Explicit CoT once backbone width reaches 124M ($d=768$; 44.96\% vs.\ 43.06\%). Official 124M GPT-2 CODI and PCCoT reproductions follow the authors' public repositories, while 19M--48M SMDM-AR baselines use our matched autoregressive implementation with identical LLaMA tokenization and training budgets.

\begin{table}[ht]
\centering
\caption{Autoregressive scaling across model capacity (19M--48M matched SlimPajama-pretrained SMDM-AR backbones and 124M GPT-2) on in-domain GSM8k ($n=1,319$), zero-shot OOD arithmetic transfer (MultiArith / SVAMP), and OOD depth extrapolation on iGSM-Small math DAGs (ID / OOD1 / OOD2). Every cell reports strict held-out unseen test accuracy (\%); best per column in each block is \textbf{bold} (ties go to StepJEPA) and second best is \textit{italic}. \dag Detached-probe refined AR runs (\cref{tab:decodability}; full-joint $w=1$ attains 28.66\% at 34M and 32.22\% at 48M; seed and selection notes in \cref{app:setup}).}
\label{tab:scaling}
\resizebox{\textwidth}{!}{%
\begin{tabular}{lcccc}
\toprule
Method (Autoregressive; Acc.\ \% $\uparrow$) & 19M & 34M & 48M & 124M (GPT-2) \\
\midrule
\multicolumn{5}{l}{\textit{GSM8k In-Domain Unseen Test Accuracy ($n=1,319$)}} \\
Direct Answer (No-CoT) & 14.56 & 16.45 & 18.88 & 17.74 \\
CODI (124M: official repo / paper) & 19.26 & 21.00 & 24.34 & 42.53 / 43.7 \\
PCCoT (124M: official repo) & 19.33 & 22.29 & 26.84 & \textbf{48.60} \\
Explicit CoT (untruncated) & \textbf{35.94} & \textbf{35.48} & \textbf{39.27} & 43.06 \\
\textcolor{ours}{\textbf{StepJEPA (refined / standard)}} & \textit{27.22} & \textit{29.87}\dag & \textit{32.15}\dag & \textit{44.96} \\
\midrule
\multicolumn{5}{l}{\textit{Zero-Shot OOD Arithmetic Transfer: MultiArith / SVAMP (trained solely on GSM8k)}} \\
Direct Answer (No-CoT) & 13.9 / 17.0 & 21.1 / 17.6 & 26.1 / 18.5 & 28.9 / 21.5 \\
CODI (124M: official repo) & 40.0 / 22.8 & 43.3 / 25.3 & 73.9 / 31.5 & 91.1 / 41.4 \\
PCCoT (124M: official repo) & 32.2 / 23.2 & 54.4 / 27.7 & 62.2 / 28.9 & \textbf{96.1} / \textbf{47.6} \\
Explicit CoT (untruncated) & \textbf{90.0} / \textbf{38.0} & \textbf{91.1} / \textbf{37.8} & \textbf{92.2} / \textbf{39.9} & 91.1 / 41.4 \\
\textcolor{ours}{\textbf{StepJEPA}} & \textit{63.9} / \textit{28.8} & \textit{75.0} / \textit{33.8}\dag & \textit{83.9} / \textit{34.0}\dag & \textit{95.0} / \textit{47.0} \\
\midrule
\multicolumn{5}{l}{\textit{iGSM-Small Math DAGs Depth Extrapolation: ID / OOD1 / OOD2 ($n=500$ each)}} \\
Explicit CoT (untruncated) & \textbf{97.2} / \textbf{30.8} / \textbf{17.6} & \textbf{96.8} / \textbf{28.0} / \textbf{15.8} & \textbf{98.0} / \textbf{26.0} / \textit{9.4} & \textbf{97.2} / \textit{55.7} / \textit{31.4} \\
CODI & 22.0 / 9.0 / 6.6 & 19.8 / 10.6 / 8.0 & 21.0 / 8.8 / 8.2 & 50.2 / 16.0 / 14.8 \\
PCCoT & 22.2 / 9.0 / \textit{8.6} & 25.8 / 10.0 / 7.4 & 21.4 / 8.2 / 7.6 & 71.4 / 23.8 / 8.6 \\
\textcolor{ours}{\textbf{StepJEPA}} & \textit{34.2} / \textit{10.4} / 6.6 & \textit{79.2} / \textit{16.0} / \textit{9.8} & \textit{86.4} / \textit{19.4} / \textbf{12.6} & \textit{89.6} / \textbf{60.2} / \textbf{35.2} \\
\bottomrule
\end{tabular}}
\end{table}

\section{Implementation Details, Hyperparameters, and Data Hygiene}
\label{app:setup}
All main StepJEPA runs (except 164M online-streamed Mazes, detailed in \cref{tab:app_hparams}) train with effective batch size 128, AdamW ($\beta_2=0.999$, gradient clipping 2.0, cosine learning-rate decay with linear warmup), dropout 0.1, $\lambda_J = 1.0$ (ramped linearly over 3,000 steps), step-decoder weight $w_{\text{dec}}=1.0$, and stopping threshold $\tau=0.5$.\footnote{We plan to release our code and models soon, pending internal approval.} Pretrained LoRA runs adapt attention and MLP projections with rank $r=128$, $\alpha=32$, and a $0.25\times$ learning-rate multiplier on the lightweight latent heads. \Cref{tab:app_arch,tab:app_hparams,tab:app_wallclock} summarize all backbone specifications, training hyperparameters, parameter overheads, and hardware-matched training wall-clock times.

\begin{table}[ht]
\centering
\caption{Strictly causal autoregressive backbone specifications and StepJEPA latent-head parameter overhead ($2d^2 + O(d)$ independent of depth).}
\label{tab:app_arch}
\small
\resizebox{\textwidth}{!}{%
\begin{tabular}{lcccccl}
\toprule
Backbone (Nominal Size) & Layers & Heads & $d$ & Pos.\ / Tokenizer & Adaptation Mode & Latent Overhead (Total / Non-Emb.) \\
\midrule
SMDM-AR 19M (non-emb.; 44--52M tot.) & 8 & 6 & 384 & RoPE / LLaMA & Scratch (full) or LoRA ($r=128$) & 0.30M--0.32M (0.59\%--0.73\% / 1.60\%--1.68\%) \\
SMDM-AR 34M (non-emb.; 78M tot.) & 8 & 8 & 512 & RoPE / LLaMA & Pretrained LoRA ($r=128$) & 0.54M (0.70\% / 1.60\%) \\
SMDM-AR 48M (non-emb.; 99M tot.) & 9 & 9 & 576 & RoPE / LLaMA & Pretrained LoRA ($r=128$) & 0.68M (0.69\% / 1.42\%) \\
GPT-2 124M (total; 145M w/ LoRA) & 12 & 12 & 768 & Learned / BPE & Pretrained LoRA ($r=128$) & 1.20M (0.83\% total / 1.41\% non-emb.) \\
AR 164M (total) & 12 & 12 & 768 & RoPE / BPE-mapped & Scratch or OpenWebText (full) & 1.31M (0.80\% / 1.54\%) \\
\bottomrule
\end{tabular}}
\end{table}

\begin{table}[ht]
\centering
\caption{Training hyperparameters across experimental regimes (all main runs use effective batch size 128, AdamW $\beta_2=0.999$, cosine decay, clip 2.0, and dropout 0.1; 164M online-streamed Mazes exceptions noted in Row 7).}
\label{tab:app_hparams}
\scriptsize
\setlength{\tabcolsep}{4pt}
\begin{tabularx}{\textwidth}{@{}l c c c c X@{}}
\toprule
Regime / Benchmark Suite & Peak LR & WD & Epochs (Warmup) & $w_{\text{cot}}$ & Latent, Rollout, and Positioning Config \\
\midrule
124M GSM8k Adaptive Seq.\ (\cref{tab:gsm124}) & $3\times 10^{-3}$ & 0.1 & 40 (3,600) & 1.0 & $K=2, M=2, S_{\text{ro}}=2$, abs.\ step emb., $J=0, \alpha_g=0, p_{\text{ro}}=0$ \\
124M GSM8k Parallel Jacobi (\cref{tab:gsm124}) & $3\times 10^{-3} \to 8\times 10^{-4}$ & 0.1 & 40 + 25 (3,600 / 1,000) & 1.0 & $K=3, I=4, H=8, w_{\text{ref}}=0.5$, 25-epoch continuation from best val \\
124M Refined OOD (iGSM / StepGame / Ded.) & $10^{-3}$ & 0.01 & 10 (234) & 1.0 & $K=2, M=2, S_{\text{ro}}=3, p_{\text{ro}}=0.35, \alpha_g=0.5, J=32$, $\mathcal{L}_{\text{SIGReg}}(\hat{z})$ (Deduction: $J=16$, CoT width 320, Jacobi $H=10$) \\
124M Refined Logic (ProsQA / StrategyQA) & $2\times 10^{-3}$ & 0.1 & 40--50 (168 / 30) & 1.0 & Refined seq.\ w/ $J=16$ (Jacobi: $K=3, I=4, H=8, w_{\text{ref}}=0.5$) \\
19M--48M Pretrained Ladder (\cref{tab:scaling}) & $10^{-3}$ & 0.1 / 0.01 & 10 (900 / 234) & 1.0 & LoRA $r=128, \alpha=32$, refined seq.\ w/ $J=16$ (GSM8k) or $J=32$ (iGSM-Small) \\
19M Scratch (\cref{tab:gen}) & $5\times 10^{-4}$ & 0.01 & 10--40 (84--234) & 0 / 1 & Full-param; ProntoQA: 40 ep, $w_{\text{cot}}=0, M=1, S_{\text{ro}}=2$; iGSM: 10 ep, refined ($w_{\text{cot}}=1, J=32$); CA1D: 10 ep, $w_{\text{cot}}=0, M=1, S_{\text{ro}}=2, J=32$; Long Mazes: 10 ep, $w_{\text{cot}}=0, M=1, S_{\text{ro}}=2, J=64$, state halting \\
164M AR Mazes (\cref{fig:main}a, \cref{tab:jepa_nec}) & $10^{-4}$ & 0.1 & 60k steps (500) & 0 & Full-param, eff.\ batch 8, AdamW ($\beta_2=0.95$), clip 1.0, dropout 0, $K=2, M=1, S_{\text{ro}}=1$ \\
\bottomrule
\end{tabularx}
\end{table}

\begin{table}[ht]
\centering
\caption{Measured training wall-clock time under matched hardware per benchmark.}
\label{tab:app_wallclock}
\small
\resizebox{\textwidth}{!}{%
\begin{tabular}{lllll}
\toprule
Benchmark & Hardware & Budget & StepJEPA ($\downarrow$) & Matched Controls and Baselines ($\downarrow$) \\
\midrule
19M ProntoQA (scratch) & 1$\times$ A100-40GB & 40 epochs & 1.10\,h & No-JEPA 1.10\,h; Explicit CoT 1.62\,h; CODI 0.85--0.87\,h; PCCoT 0.63--0.64\,h \\
124M iGSM-Small & 1$\times$ H200 & 10 epochs & 8.07\,h & No-JEPA 8.13\,h \\
124M GSM8k (LoRA) & 4$\times$ A100-40GB DDP & 40 / 40+25 epochs & 65.1\,h ($\approx$260 GPU-h) Seq / 84.3\,h ($\approx$337 GPU-h) Jacobi & -- (official repo baselines run separately) \\
\bottomrule
\end{tabular}}
\end{table}

\textbf{Inference accounting, latency, and checkpoint selection.}
\begin{itemize}
  \item \emph{NFE, emitted tokens, and latency (\cref{tab:gsm124}):} NFE counts true per-example sequential forward passes at batch size 1 across all $n=1,319$ GSM8k problems for StepJEPA, Direct Answer, and student-only official PCCoT, and across the $n=300$ latency subset for Explicit CoT and fused-pass KV-cached CODI. Emitted counts greedy BPE subword tokens including the end-of-sequence token uniformly across methods (2.24 adaptive StepJEPA, 2.22 Jacobi StepJEPA, 2.2 PCCoT, 6.2 Direct Answer, 6.3 CODI, 31.3 Explicit CoT). Same-GPU latency benchmarks all models on one NVIDIA A100-40GB ($n=300$, merged LoRA) in fp32; bf16 preserves the identical ranking (223.6\,ms CoT, 107.7\,ms CODI, 78.5\,ms StepJEPA Seq, 59.3\,ms StepJEPA Jacobi, 52.1\,ms PCCoT), with StepJEPA uncached while baselines use standard KV caching.
  \item \emph{Baseline selection and seed notes (\cref{tab:gen,tab:scaling}):} 124M PCCoT uses strict validation-selected learning rate $3\times 10^{-3}$ on StepGame (99.0\% val vs.\ 97.0\% at $10^{-3}$, which attains 93.6 / 83.8 / 52.6, still $-37.4$ pts below StepJEPA on 10-hop OOD) and iGSM-Small (67.0\% vs.\ 55.7\% val at $10^{-3}$), and trains 30 epochs on ProsQA (validation peaking at epoch 24, step 3,360; ProsQA CODI trains 40 epochs at matching learning rate $3\times 10^{-3}$ with final-step dropping, peaking at step 5,180). On 19M from-scratch Long Mazes, No-JEPA remained at 0.0\% validation accuracy through epoch 8 and a 2-epoch continuation at learning rate $10^{-4}$ (10.0 epochs total), and CODI/PCCoT use the matched unit-distillation-weight from-scratch protocol consistent with 19M CA1D and iGSM-Small. On 124M Multi-Hop Deduction, StepJEPA completed its final 2 epochs via a continuation from epoch 8 at learning rate $1.5\times 10^{-4}$ (10.0 epochs total; epoch 8 attained 100.0 / 68.8 / 48.0), No-JEPA reports its epoch-5 checkpoint (step 3,910, 87.0\% validation accuracy vs.\ StepJEPA's 100.0\% by epoch 5), and Explicit CoT is evaluated with an unclipped 700-token generation budget up to GPT-2's 1,024-token context limit (where 71.6\% of OOD2 verbal chains, 110/252 9-hop and 248/248 10-hop, exceed 1,024 tokens, leaving 28.4\% within context vs.\ 28.8\% accuracy). In \cref{tab:scaling}, 19M--48M Explicit CoT uses an untruncated 256-token generation budget (seed 0; 19M 3-seed mean: 36.04\% GSM8k and 91.1 / 38.2 MultiArith / SVAMP); 124M Explicit CoT reports exact $n=1,319$ GSM8k accuracy (43.06\%) and untruncated OOD from the same checkpoint, and the 3-seed mean on iGSM-Small (97.2 / 55.7 / 31.4; seeds 0--2: 97.0/60.4/30.4, 97.8/60.6/35.6, 96.8/46.2/28.2); 19M StepJEPA reports the dedicated ladder run (27.22\%, matching the 26.69\% ablation baseline in \cref{tab:decodability,tab:app_v3} within run-to-run variance).
\end{itemize}

\textbf{Data and evaluator hygiene audit.} Every benchmark underwent an automated quality-assurance audit verifying: (i) zero exact or normalized problem overlaps across train, validation, and test splits; (ii) 100\% self-consistency of gold-answer verifiers; (iii) zero test-set truncation or example dropping across all latent-inference splits (with minor training-only length filtering of 9/385,620 GPT-2 GSM8k, 170/385,620 LLaMA GSM8k, and 95/1,609 train and 8/200 val StrategyQA-CoT examples exceeding single-step width, auxiliary training-only CoT width capped at 128 or 320 tokens, and GPT-2's architectural 1,024-token context ceiling on 71.6\% of 124M Deduction Explicit CoT OOD2); and (iv) strict test-set isolation, with all checkpoint selection performed solely on validation splits.

\section{Negative Results and Shortcut Audits}
\label{app:neg}
\textbf{Negative results.} (i) Training a reinforcement-learning router via REINFORCE \citep{williams1992reinforce} to alternate between writing text and thinking latently traced a modest accuracy--token Pareto frontier on graph tasks without improving out-of-distribution depth extrapolation. (ii) On StepGame, our shortcut audit using a separate 170M bidirectional answer-only diagnostic model (trained without intermediate reasoning steps) revealed that direct answer prediction attains 99.6\% ID (99.8\% 4-hop), 98.3\% 8-hop OOD1, 72.8\% 10-hop OOD, and 81.1\% 9--10-hop OOD2 via prompt cues; we report this transparently and center our main conclusions on shortcut-free benchmarks (GSM8k, ProsQA, ProntoQA, Mazes, iGSM-Small, 1D Cellular Automata, and Multi-Hop Deduction).

\end{document}